\documentclass[10pt,journal,compsoc]{IEEEtran}
\ifCLASSOPTIONcompsoc
  \usepackage[nocompress]{cite}
\else
  \usepackage{cite}
\fi

\ifCLASSINFOpdf
   \usepackage[pdftex]{graphicx}
   \DeclareGraphicsExtensions{.pdf,.jpeg,.png}
\else
   \usepackage[dvips]{graphicx}
   \DeclareGraphicsExtensions{.eps}
\fi

\usepackage{subfigure}
\usepackage{float}
\usepackage[cmex10]{amsmath}
\usepackage{amsfonts}
\usepackage{mathtools}
\usepackage{multirow}
\usepackage{graphicx}
\usepackage{algpseudocode}
\usepackage{algpascal}
\usepackage{algc}
\usepackage{array}
\usepackage{float}
\usepackage{bm}
\usepackage{booktabs}
\usepackage{multirow}
\usepackage{url}
\usepackage{color}

\usepackage{algorithm}

\begin{document}
\title{Multi-modal Multi-kernel Graph Learning for Autism Prediction and Biomarker Discovery}

\author{
        Jin Liu$^{*}$, 
        Junbin Mao, 
        Hanhe Lin, 
        Hulin Kuang, 
        Shirui Pan, 
        Xusheng Wu$^{*}$,
        Shan Xie, 
        Fei Liu, 
        Yi Pan$^{*}$
\IEEEcompsocitemizethanks{

\IEEEcompsocthanksitem Jin Liu is with the 1) Hunan Provincial Key Lab on Bioinformatics, School of Computer Science and Engineering, Central South University, Changsha 410083, China; 2) Xinjiang Engineering Research Center of Big Data  and Intelligent Software, School of software, Xinjiang University, Wulumuqi 830000, China.
EMAIL: liujin06@csu.edu.cn

\IEEEcompsocthanksitem Junbin Mao, Hulin Kuang are with the Hunan Province Key Lab on Bioinformatics, School of Computer Science and Engineering, Central South University, Changsha 410083, China.
EMAIL: maojunbin@csu.edu.cn; hulinkuang@csu.edu.cn

\IEEEcompsocthanksitem Yi Pan is with the 1) Faculty of Computer Science and Control Engineering,
Shenzhen University of Advanced Technology; 2) Shenzhen Institute of Advanced Technology, Chinese Academy of Sciences, Shenzhen 518055, China.
EMAIL: yi.pan@siat.ac.cn

\IEEEcompsocthanksitem Hanhe Lin is with the School of Science and Engineering, University of Dundee, DD1 4HN Dundee, United Kingdom.
EMAIL: hlin001@dundee.ac.uk

\IEEEcompsocthanksitem Shirui Pan is with the School of Information and Communication Technology, Griffith University, Gold Coast, QLD 4215, Australia.
EMAIL: s.pan@griffith.edu.au

\IEEEcompsocthanksitem Xusheng Wu, Shan Xie, and Fei Liu are with the Shenzhen Health Development Research and Data Management Center, Shenzhen 518109, China.
EMAIL: jysgwuxusheng@163.com; szxieshan@126.com; liufei\_csu@163.com

\IEEEcompsocthanksitem $^{*}$ Corresponding author}
}

\IEEEtitleabstractindextext{%
\begin{abstract}
Graph learning-based multi-modal integration and classification is one of the most challenging tasks for disease prediction. To effectively offset the negative impact among modalities in the process of multi-modal integration and heterogeneous information extractions from graphs, we propose a novel method called Multi-modal Multi-Kernel Graph Learning (MMKGL). To solve the problem of negative impact among modalities, we propose a multi-modal graph embedding module to construct a multi-modal graph. Different from conventional methods that manually construct static graphs for all modalities, each modality generates a separate graph by adaptive learning, where a function graph and a supervision graph are introduced for optimization during the multi-graph fusion embedding process. We then propose a multi-kernel graph learning module to extract heterogeneous information from the multi-modal graph. The information in the multi-modal graph at different levels is aggregated by convolutional kernels with different receptive field sizes, followed by generating a cross-kernel discovery tensor for disease prediction. Our method is evaluated on the benchmark Autism Brain Imaging Data Exchange (ABIDE) dataset and outperforms the state-of-the-art methods. In addition, discriminative brain regions associated with autism are identified by our model, providing guidance for the study of autism pathology. The source code will be available at \url{https://github.com/yutian0315/MMKGL}.
\end{abstract}

}

\maketitle

\IEEEdisplaynontitleabstractindextext

\IEEEpeerreviewmaketitle


\section{Introduction}
\IEEEPARstart{T}{he} development of medical devices has facilitated the acquisition of multi-modal clinical data. It has been found that multi-modal data could benefit disease prediction as it contains more information than single-modal data~\cite{wang2022machine, lv2024disentangled} by taking advantage of the joint and supplementary information. Recently, a few approaches that aim to make use of multi-modal data for disease prediction have been proposed, including multi-modal fusion methods based on attention mechanisms \cite{10325596, zheng2022multi}, multi-task learning \cite{adeli2019multi, liu2024multi}, and nonnegative matrix decomposition \cite{zhou2019latent, peng2020group}. However, these approaches rely on feature learning and do not adequately explore the structure and cross-modal relationships between modalities. Simple modal fusion cannot fully utilize multi-modal data because different modalities lead to complex relational dependencies. Graph learning works by connecting data from different modalities into edges in an optimal graph and constructing specific learning paradigms for various downstream tasks\cite{liu2022mmgk, 10746501, wu2023medical}. As a result, how to fully exploit the knowledge of multi-modal is still a long-standing research topic. 

Owing to the inherent properties of graphs \cite{su2020network} such as information aggregation and relationship modeling, graph models provides an efficient way for integrating multi-modal information. Graph neural networks (GNNs) have been proposed in \cite{welling2016semi}. Due to their advantages of joint population learning and information mining, a growing number of GNN-based studies have been proposed for disease prediction. PopGCN \cite{parisot2017spectral} is the first application of graph convolutional networks to Autism and Alzheimer's disease. It combines brain imaging information and phenotypic information jointly to construct disease population graphs and perform disease prediction to distinguish normal individuals from patients. InceptionGCN \cite{kazi2019inceptiongcn} explored the impacts of the size of convolutional filters on disease prediction based on PopGCN, and tried to find the optimal convolutional filter size. MMGL \cite{zheng2022multi} used a self-attentive mechanism to learn complementary information between modalities and is used to construct disease population graphs. RAGNN \cite{10373946} proposed a regionally asymmetric adaptive graph convolutional network to learn non-Euclidean spatial features of the left and right hemispheres of the brains of autistic patients. WL-deepGCN\cite{10247643} utilized a weight learning network to construct population graph edge weights for autism diagnosis. 

Although the above studies have achieved promising results for disease prediction, there are some limitations. One of the most challenging problems is that the domain distributions of different modalities vary significantly. It leads to the fact that when multi-modal data are directly integrated, the less expressive modalities may suppress the expression of other modal data. This phenomenon is known as the inter-modal negative impact. Another major issue is that most of the existing graph convolutional network-based disease prediction methods use a size-fixed convolutional filter, which cannot well extract the heterogeneous information over the graph.

To overcome these problems and to effectively utilize multimodal data, we propose a novel framework entitled Multi-modal Multi-Kernel Graph Learning (MMKGL). The contributions of our work are summarized as follows:

\begin{itemize}
    \item We propose a multi-modal graph embedding module that generates multiple graphs adaptively, where each graph corresponds to one modal data. A function graph and a supervision graph are introduced to optimize the multi-graph fusion embedding process, which effectively mitigates the negative impacts between modalities.

    \item We propose a multi-kernel graph learning network that extracts heterogeneous information by using convolutional kernels with different receptive field sizes. To minimize the graph noise, a relational attention mechanism (RAM) is deployed to adaptively tune the graph structure in the training process.
b n
    \item The proposed framework is evaluated on the Autism Brain Imaging Data Exchange (ABIDE) dataset. The results verify the validity of our proposed framework and demonstrate its advantages over state-of-the-art methods. 

    \item Discriminative brain regions and subnetworks with discriminatory properties are discovered, giving guidance for the study of autism pathology.
\end{itemize}

\section{Related Work}
\subsection{Multi-modal Analysis for Disease Prediction}
Multi-modal analysis for disease prediction aims to explore complementary and specific information of modalities in a specific way and use them for disease prediction. Traditional multi-modal approaches usually concatenate multi-modal features in the original space to perform disease prediction. However, such approach cannot take full advantage of the complementary information present in multi-modal data.

An increasing number of researchers have started to explore more intricate multi-modal approaches\cite{peng2020group,zhou2019latent,zheng2022multi,song2022multi,9740146,9857948}. Peng \emph{et al}.\cite{peng2020group} utilized a joint non-negative matrix decomposition algorithm (GJNMFO) to identify abnormal brain regions associated with diseases by projecting three modal data into different coefficient matrices. Zhou \emph{et al}.\cite{zhou2019latent} learned a latent space for multi-modal data that retains modality-specific information, and then projected the features of the latent space into the label space for prediction. Xue \emph{et al}.\cite{song2022multi} constructed a fused brain functional connectivity networks by applying different strength penalty terms for different modalities. To combine heterogeneous structural information between multi-modal states, Zhang \emph{et al}.\cite{9740146} proposed a fusion algorithm based on graph popular regularization. Zhu \emph{et al}.\cite{9857948} proposed a triple network to explore higher-order discriminative information in multi-modal while extracting complementary information from fMRI and DTI (Diffusion Tensor Imaging) using cross-attention. Zheng \emph{et al}.\cite{zheng2022multi} exploited attention-based pattern-aware learning to mine complementary information between multi-modal data. Unlike the fusion at the feature level in \cite{zheng2022multi}, our framework transforms multi-modal into a graph for fusion. 

\subsection{Graph Neural Networks in Disease Prediction}
Graph neural networks (GNNs) provide a practical solution for exploiting potential information from different modalities.
Most early studies on GNNs-based disease prediction are based on either single modal data \cite{kazi2019inceptiongcn} or manually constructed static graphs \cite{parisot2017spectral}. For the former, it lacks rich multi-modal information. For the latter, it brings a lot of noise to the graph.

Some GNNs methods using multi-modal data and adaptive graph learning methods have been proposed in recent years~\cite{huang2022disease,cosmo2020latent,zheng2022multi,9770779,9766117,zhang2022classification}. Huang \emph{et al}.\cite{huang2022disease} constructed a pairwise encoder between subjects from phenotypic information to learn the adjacency matrix weights. Cosmo \emph{et al}.\cite{cosmo2020latent} learned optimal graphs for downstream classification of diseases based on dynamic and local graph pruning. Ji \emph{et al}.\cite{9770779} learned graph hash functions based on phenotypic information and then used them to transform deep features into hash codes to predict the classes of brain networks. Li \emph{et al}.\cite{9766117} proposed an edge-based graph path convolution algorithm that aggregates information from different paths and such algorithm is suitable for dense graphs like brain function networks. Hao \emph{et al}.\cite{zhang2022classification} designed a local-to-global graph neural network to explore the global relationships between individuals and the local connections between brain regions. Zheng \emph{et al}.\cite{zheng2022multi} used the complementary and shared information between multi-modalities to learn the structure of graphs through an adaptive learning mechanism of graphs. Different from the two-layer graph convolutional network used in \cite{zheng2022multi}, the multi-kernel graph learning network of our framework contains different convolutional kernels that can efficiently extract the heterogeneous information of a graph.

\begin{figure*}[htbp]   
	\centering
	\includegraphics[width=\linewidth,scale=1.00]{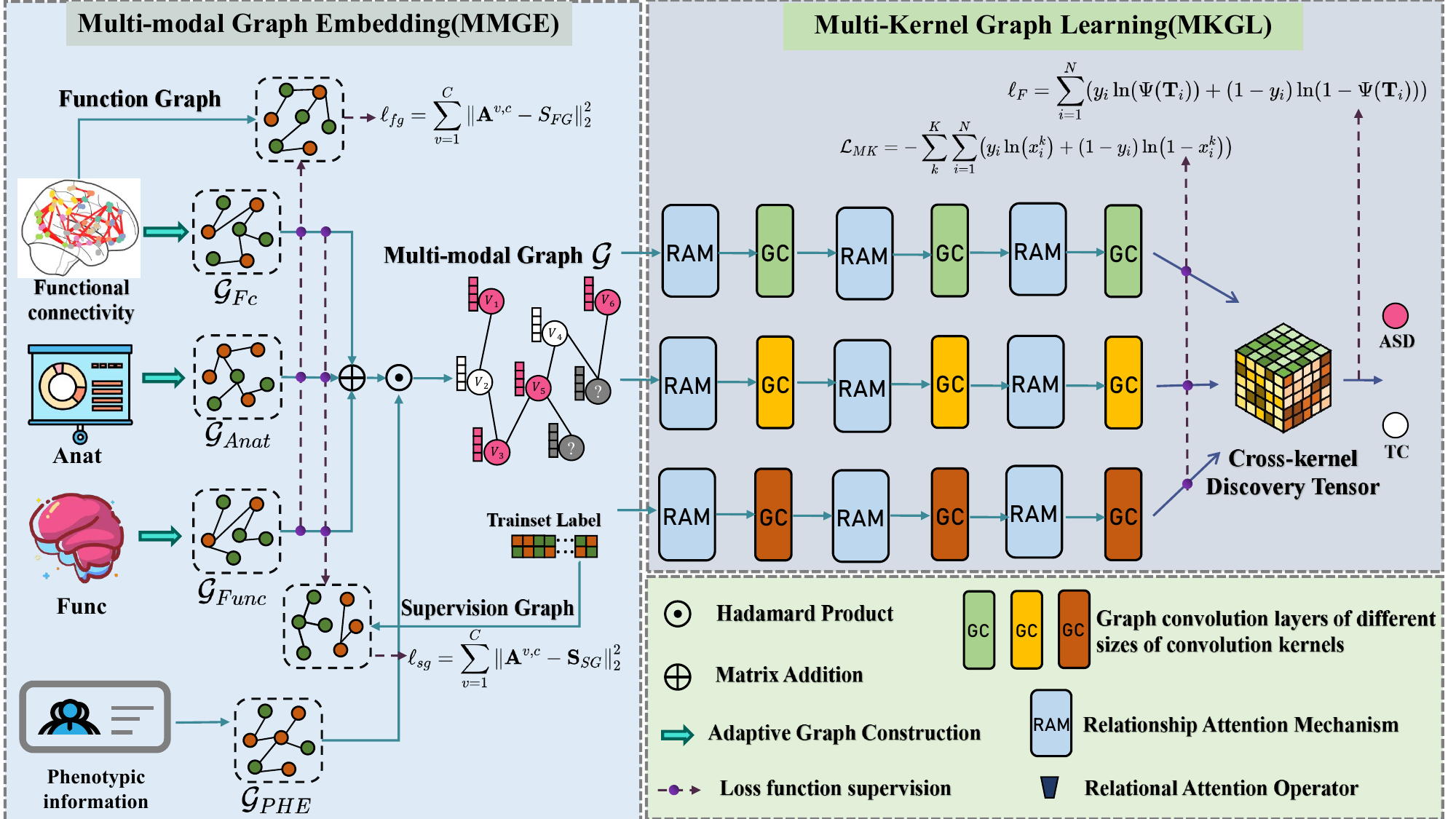}
	
	\caption{Overview of the architecture of our framework. \textbf{Multi-modal graph embedding}: Multiple modal graphs $\mathcal{G}_{fc}$, $\mathcal{G}_{anat}$, $\mathcal{G}_{func}$ are generated by adaptive graph construction and then are fused into a Multi-modal graph $\mathcal{G}$. The function graph $S_{FG}$ and supervision graph $S_{SG}$ are introduced in the fusion process for optimization. \textbf{Multi-kernel graph learning}: The heterogeneous information of the Multi-modal graph $\mathcal{G}$ is extracted using a graph convolutional network with different convolutional kernel sizes and fused by generating a cross-kernel discovery tensor \textbf{T} for predicting autism spectrum disordor (ASD) and typical control (TC). In addition, relationship attention mechanism (RAM) is used to adjust the weights of the adjacency matrix \textbf{A}.}
	\label{fig:framework}
\end{figure*}

\section{METHODOLOGY}
\subsection{Overview}

For multi-modal data, the inputs of the \emph{m}-th modality consists of features $ \textbf{X}^{m} = \left\{x_{1}^{m}  , x_{2}^{m} , \ldots, x_{N}^{m} \right\} $ of \emph{N} subjects. For the multi-modal features of each subject $x_{i} \in  \textbf{X}$, there is an associated label $y_{i} \in  \textbf{Y}$. The task is defined on the dataset with \textbf{X} as the input and \textbf{Y} as the prediction target. Given a graph $ \mathcal{G}=\left (  \mathcal{V}, \mathcal{E}, \textbf{X}\right )$, $\mathcal{V}$ is a set of nodes and $\mathcal{E}$ is a set of edges (e.g. $e_{ij} \in \mathcal{E}$ if and only if the distance between $v_{i}$ and $v_{j}$ is below a threshold), where $v_{i} \in \mathcal{V}$ denotes the $i$-th node which represents a subject in autism prediction. $e_{ij} = \langle v_{i},v_{j} \rangle $ describes the relationship between subject $i$ and $j$. In addition, the adjacency matrix $\textbf{A} \in \mathbb{R}^{N \times N}$ is derived from the set of edges $\mathcal{E}$, where $\textbf{A}_{ij}$ corresponds to the weights of $e_{ij}$.

As shown in Fig.~\ref{fig:framework}, our proposed framework mainly consists of two modules, i.e., Multi-modal Graph Embedding and Multi-Kernel Graph Learning. 

\textbf{Multi-modal Graph Embedding} constructs a separate graph for each modality by adaptive graph construction. Then, multi-graph fusion is performed under the supervision of a function graph $S_{FG}$ and a supervision graph $S_{SG}$. Finally, phenotypic information is embedded into the fused graph to generate a multi-modal graph $\mathcal{G}$.

\textbf{Multi-Kernel Graph Learning} acquires heterogeneous information of the multi-modal graph $\mathcal{G}$ through convolutional kernels of different sizes. Among them, relational attention mechanism is proposed to adjust the weights of individual relations in the multi-modal graph $\mathcal{G}$ with the self-adaptation. Finally, a cross-kernel discovery tensor is generated by fusing the heterogeneous information and used for autism prediction.



\subsection{Data acquisition and Preprocessing}

We validate our framework on the publicly available Autism Brain Imaging Data Exchange (ABIDE) \cite{ADiMartino2014TheAB} dataset. It collects data from 17 international sites and contains neuroimaging and phenotypic data of 1,112 subjects. 
In this study, followed by the same imaging criteria in \cite{abraham2017deriving}, 871 subjects were selected, where 403 are with ASD and 468 are with typical control (TC). To ensure equitability and reproducibility, the fMRI data was processed by using the Configurable Pipeline for the Analysis of Connectomes (CPAC) from the Preprocessed Connectomes Project (PCP) \cite{craddock2013neuro}. The preprocessed functional images were registered to a standard anatomical space (MNI152). Subsequently, the average time series of brain regions were extracted according to the automatic anatomical atlas (AAL).

Four modalities were selected for this study. 1) \emph{Brain functional connectivity} (FC): Calculated from time series of brain regions based on Pearson's correlation coefficient. 2) \emph{Phenotype information} (PHE): Including age, gender, site, and scale information. 3) \emph{Automated anatomical quality assessment metrics} (Anat): Reflecting indicators such as smoothness of Voxels, percentage of Artifact Voxels, and signal to noise ratio. 4) \emph{Automated functional quality assessment metrics} (Func): Reflecting indicators such as entopy focus criterion, standardized DVARS, and mean distance to median volume. Anat and Func are proven to be effective in disease prediction \cite{zheng2022multi}.

\subsection{Multi-modal Graph Embedding}
\subsubsection{Graph Construction}
We divide the four modalities into two categories based on their properties: 1) Continuous Data $\textbf{X}^{c}$: FC, Anat, and Func. 2) Discrete Data $\textbf{X}^{d}$: PHE.

For continuous features $\textbf{X}^{c}$, we calculate cosine similarity between pairs of subjects and maps onto the [0, 1] interval by rescaling to construct an adjacency matrix $\textbf{A}^{v,c}$ of $v$-th modality, where $v$ includes FC, Anat, and Func.  
The similarity between subject $i$ and subject $j$ is defined as:
\begin{equation}
\textbf{A}_{ij}^{v,c} = \frac{(x_{i}^{v}\Theta^{v})^{\top} x_{j}^{v}\Theta^{v}}{2\left\|x_{i}^{v}\Theta^{v}\right\|\left\|x_{j}^{v}\Theta^{v}\right\|}+0.5,
\label{eq}
\end{equation}
where $x_{i}^{v}$ is the feature of $v$-th modality of $i$-th subject. In order to allow the features of each modality to learn adaptively during the composition, the projection transformation of the subspace is applied to the features before estimating the cosine similarity. $\Theta^{v}$ is the projection matrix (Implemented by a full connectivity layer) of the $v$-th modal.

For discrete features $\textbf{X}^{d}$, i.e., PHE, we calculate the correlation between pairs of subjects using attribute matching. We construct an adjacency matrix $\textbf{A}^{v,d}$ for $v$-th modality. 
The similarity between subject $i$ and subject $j$ is estimated as:

\begin{equation}
\textbf{A}_{ij}^{v,d} =\sum_{t=1}^{T} \phi _{t} \left ( P_{i}^{v,t} , P_{j}^{v,t} \right ), \label{eq}
\end{equation}
where $\phi _{t}$ denotes the attribute matching function used in discrete features for variables of type $t$, $P$ is the discrete feature of its corresponding type. For example, if $P$ is gender with a binary type feature, we let $\phi $ to be $\left \{ e_{ij} = 1 \parallel if\ P_{i} =P_{j}   \right \} $. Similarly, when $P$ is age data, we let $\phi $ to be $\left \{ e_{ij} = 1 \parallel if\ |P_{i} -P_{j}|<\theta \right \}$.

In addition, we introduce a supervision graph $\mathbf{S}_{SG}$ and a function graph $\mathbf{S}_{FG}$ for optimization during the multi-graph fusion embedding, defined as:

\begin{equation}
\mathbf{S}_{SG}^{i j}=\left\{\begin{array}{lll}
1, & y_{i}=y_{j} &  { (train \ labels) } \\
0, & y_{i} \neq y_{j} &  { (train \ labels) }
\end{array}\right.,
\end{equation}

\begin{equation}
\label{eq2}
\mathbf{S}_{FG}^{i j} =\emph{exp} \left(-\frac{\left(\mathbf{x }_{i}^{FC}-\mathbf{x }_{j}^{FC}\right)^{2}}{2 \sigma^{2}}\right),
\end{equation}

\noindent where $\sigma$ is Gaussian kernel, $\mathbf{x }_{i}^{FC}$ and $\mathbf{x }_{j}^{FC}$ are FC features of subject $i$ and $j$. The reason for choosing $FC$ is that it has the best representational ability.

\subsubsection{Graph Integration}

In order to learn towards the optimal graph for $\textbf{A}^{c}$, we design a loss function to optimize the learning of graph structure. The reason for constructing $\mathbf{S}_{SG}$ is that GNNs are able to classify unknown nodes quite simply if given a fully labeled graph (optimal graph). The loss is defined as follows:

\begin{equation}
\ell _{sg} =\sum_{v=1}^{C}\left \| \textbf{A}^{v,c} - \mathbf{S}_{SG} \right \|_{2}^{2},
\label{eq4}\end{equation}

\noindent where $\mathbf{S}_{SG}$ is the supervision graph. We supervise the construction of the modal graph by $\mathbf{S}_{SG}$ with the expectation that the transformation ability of the projection matrix of $\Theta$ will have good generalization ability.

Meanwhile, to prevent the overfitting problem caused by the supervision graph $\mathbf{S}_{SG}$, we introduce a function graph $\mathbf{S}_{FG}$ constructed by the modal feature with the best expressiveness. Function graph $\mathbf{S}_{FG}$ supervise the generation of validation and test set nodes as a way to enhance the generalization of the model, defined as follows:

\begin{equation}
\ell _{fg} =\sum_{v=1}^{C}\left \| \textbf{A}^{v,c} - \mathbf{S}_{FG} \right \|_{2}^{2}.
\label{eq5}\end{equation}

Finally, the objective function of Multi-modal Graph Embedding can be expressed as the following equation:

\begin{equation}\mathcal{L}_{MMGE} =\ell _{sg}+ \ell _{fg}.\label{eq}\end{equation}

\noindent  Using the loss function $\mathcal{L}_{MMGE}$, we can optimize the generation and fusion embedding of multiple modal graphs well to obtain a multi-modal graph $\mathcal{G}$ that is close to the optimal graph and use it for autism prediction.

The adjacency matrix \textbf{A} of the multi-modal graph $\mathcal{G}$ can be obtained from the adjacency matrices $\textbf{A}^{v,c}$ and $\textbf{A}^{v,d}$, which is given by:

\begin{equation}
\textbf{A} = \frac{1}{C}\sum_{v=1}^{C} \textbf{A}^{v,c} \odot\frac{1}{D}  \sum_{v=1}^{D} \textbf{A}^{v,d},
\end{equation}

\noindent where $C$ and $D$ are the number of continuous and discrete modalities, respectively.  
$\odot$ is matrix Hadamard product. In our experiment, $C$ and $D$ are 3 and 1, respectively. 

\subsection{Multi-Kernel Graph Learning}

The multi-modal graph $\mathcal{G}$ consists of edges $\mathcal{E}$ (relationships) and nodes $\mathcal{V}$ (subjects). To characterize the connections between nodes and nodes in the multi-modal graph $\mathcal{G}$, a better choice is the regularized Laplacian matrix: $L=I-D^{-\frac{1}{2}} \textbf{A} D^{-\frac{1}{2}}$, where \textbf{A} is the multi-modal graph adjacency matrix, $D$ is the degree matrix of the nodes, and $I$ is a identity matrix. Since $L$ is a real symmetric positive semdefinite matrix, we can use the matrix decomposition as $L=U \Lambda U^{\mathrm{T}}$, where $U$ is a matrix composed of eigenvectors and $\Lambda$ is a matrix composed of eigenvalues. For the node features \textbf{X} ($\textbf{X}^{FC}$ is replaced by \textbf{X} for simplicity of writing) and the adjacency matrix \textbf{A}, we can obtain the representation of the graph convolution $\Phi(\mathcal{G})$ on the multi-modal graph $\mathcal{G}$:

\begin{equation}
\Phi(\mathcal{G}) = \Phi(\textbf{X} * \textbf{A})=U \textbf{A}_{\theta} U^{T} \textbf{X},
\end{equation}
\noindent where $\textbf{A}_{\theta}=diag \left(U^{T} \textbf{A}\right)$. To reduce calculation costs $O\left(N^{2}\right)$, Chebyshev graph convolution uses Chebyshev polynomials to approximate the spectral graph convolution. In the field of polynomial function approximation, Chebyshev polynomials are usually preferred due to their numerical stability and computational efficiency. Introducing the polynomial, let $\textbf{A}_{\theta}=\sum_{i=0}^{k} \theta_{i} \Lambda^{i}, \theta \in R^{k}$, so that the following equation is obtained:

\begin{equation}
\Phi(\mathcal{G}) = \Phi(\textbf{X} * \textbf{A})=U \sum_{i=0}^{k} \theta_{i} \Lambda^{i} U^{T} \textbf{X}.
\end{equation}

Then by shifting the eigenvector $U$ right into the summation equation $\sum_{i=0}^{k} \theta_{i} \Lambda^{i} U^{T} \textbf{X}$ and passing the equation $L=U \Lambda U^{\mathrm{T}}$, we can obtain the following equation:

\begin{equation}
\label{eq1}
\Phi(\mathcal{G}) = \Phi(\textbf{X} * \textbf{A})=\sum_{i=0}^{k} \theta_{i} L^{i} \textbf{X},
\end{equation}

\noindent where $k$ is the order of the Chebyshev polynomial. $T_{k}(L)=2 L T_{k-1}(L)-T_{k-2}(L)$ is the Chebyshev polynomial defined recursively with $T_{0}(L) = 1$ and $T_{1}(L)=L$. Bringing it into Eq.~(\ref{eq1}) shows that:

\begin{equation}
\Phi(\mathcal{G}) = \Phi(\textbf{X} * \textbf{A})=\sum_{i=0}^{k} \theta_{i} T_{i}(\tilde{L}) \textbf{X},
\end{equation}

\noindent where $ \tilde{L}=\frac{2 L}{\lambda_{\max }}-I$, $\tilde{L}$ is the rescaled graph Laplace operator. Similar to convolutional neural networks, the polynomial $T_{k}(\tilde{L})$ is a $K$-order domain aggregator that integrates the information of neighboring nodes at $K$ steps from the central node.

According to the above derivation, the forward propagation form of the Chebyshev convolutional layer can be obtained:

\begin{equation}
\Phi_{k}(\mathcal{G})  =\Phi_{k}(\textbf{X} * \textbf{A})=\sum_{i=0}^{k} T_{i}(\tilde{L}) \textbf{X}  W_{i},
\end{equation}

\noindent where $\Phi_{k}(\mathcal{G})\in \mathbb{R}^{N \times H'}$ represents the Chebyshev convolution of order $k$ on the graph $\mathcal{G}$. $\textbf{X}  \in \mathbb{R}^{N \times H}$, where $H$ is the feature dimension of input \textbf{X}. $W_{i}  \in \mathbb{R}^{H \times H'}$, where $H'$ is the output dimension of the fully connected layer $W_{i}$.

\subsubsection{Cross-kernel Discovery Tensor}
With the information aggregation of multiple Chebyshev graph convolutional networks of different orders, we obtain the heterogeneous information of the input features \textbf{X} on the graph $\mathcal{G}$. Before \textbf{X} is input to the feature fusion module, we train each Chebyshev network by the cross-entropy loss function so that its output $\Phi_{k}(\mathcal{G})$ has a better representation before fusion. The loss function is shown as follows:

\begin{equation}
\ell _{MK}=-\sum_{k}^{K} \sum_{i=1}^{N}\left(y_{i}  \ln \left(\hat{x} _{i}^{k} \right)+\left(1-y_{i}\right)  \ln \left(1-\hat{x} _{i}^{k}\right)\right),
\end{equation}

\noindent where $\hat{x} _{i}^{k}$ is the output of a Chebyshev convolutional network of order $k$. Next, we use the fusion module to fuse the outputs of multiple Chebyshev convolutional networks. Fusion module is designed to learn the cross-correlation of higher-level intra-view and heterogeneous graph information in the label space. For the predicting probability distribution ($\hat{x}_{i}^{(k)}, k=\left \{  1,2,\dots ,K\right \}$ )  of the $i$-th sample from the output of different Chebyshev convolutional networks $\Phi_{k}$, we construct a Cross-Kernel Discovery Tensor (CKDT) $\mathbf{T}_{i} \in \mathbb{R}^{t^{K}}$, where $t$ is the number of classes. The formula is defined as follows:

\begin{equation}
\mathbf{T}_{i}=\prod_{k=1}^{K} \hat{x}_{i}^{(k)} , k=1,2, \ldots, K
\end{equation}

$\mathbf{T}_{i}$ is then flattened to a 1-dimensional vector and the final prediction is made using the fully connected network $\Psi \left (  \cdot \right ) $, where the loss function is written as:

\begin{equation}
\ell _{F}=\sum_{i=1}^{N}  \left(y_{i} \ln \left(\Psi \left (\mathbf{T}_i  \right ) \right)+\left(1-y_{i}\right) \ln \left(1-\Psi \left (\mathbf{T}_i  \right )\right)\right).
\end{equation}

The objective function of Multi-Kernel Graph Learning can be expressed as the following equation:
\begin{equation}
\mathcal{L}_{MKGL} =\ell _{MK}+ \ell _{F}.\label{eq}
\end{equation}

Eventually we optimize our model using $\mathcal{L}_{MMGE}$ and $\mathcal{L}_{MKGL}$ until convergence, and the total loss function can be expressed as follows:

\begin{equation}
\mathcal{L}=\lambda_{1}  \mathcal{L}_{MMGE}+\lambda_{2}\mathcal{L}_{MKGL}
\end{equation}

\noindent where $\lambda_{1}$ and $\lambda_{2}$ are the weight parameters of the corresponding loss functions, respectively. Algorithm~\ref{alg:1} details the procedure of our proposed MMKGL framework.  

\subsubsection{Relational Attention Mechanism}
To reduce the noise of the multi-modal graph $\mathcal{G}$. we propose a Relational Attention Mechanism (RAM) to learn specific information between subjects. Specifically, we first filter the more valuable individual relationships from the multi-modal graph $\mathcal{G}$ by threshold. A less noisy adjacency matrix $\hat{\textbf{A}}$ is generated by the learnable parameters $\psi$, i.e., edges of the same class are weighted more and those across class are weighted less. The RAM can be expressed as:

\begin{equation}
R_{i j}=\xi \left(\psi  \theta_{i}, \psi  \theta_{j}\right),
\end{equation}

\noindent where $\theta_{i}, \theta_{j}$ are the subject's FC feature embedding, and $R_{i j}$ is the learned relational attention score, which represents the informational relational reference importance of subject $j$ to subject $i$. Learned parameter $\psi  \in \mathbb{R} ^{F\times Z }$, where $F$ is the dimension of feature $\theta$ and $Z$ is the dimension of the hidden unit, $\xi  \in \mathbb{R} ^{2Z\times 1 } $ is the relational attention operator, To make the relational attention scores easily comparable across subjects, we normalize all choices for subject $j$ using the \emph{Softmax} function:
\begin{equation}
 \hat{\textbf{A}}_{i j}=\emph{Softmax} \left(R_{i j}\right)=\frac{\exp \left(R_{i j}\right)}{\sum_{e \in N_{i}} \exp \left(R_{i e}\right)},
\end{equation}

\noindent where $ \hat{\textbf{A}}_{i j}$ is the normalize relational attention weight and $N_{i}$ is the neighboring node with which subject $i$ is associated. 

In this study, the relational attention operator $\xi$ is a single-layer feedforward neural network. The information relationship between subjects can be expressed as:

\begin{equation}
 \hat{\textbf{A}}_{i j}=\frac{\exp \left( \sigma \left(\xi^{T}\left[\psi \theta_{i} \sqcup  \psi \theta_{j}\right]\right)\right)}{\sum_{e \in N_{i}} \exp \left( \sigma \left(\xi^{T}\left[\psi \theta_{i} \sqcup  \psi \theta_{e}\right]\right)\right.},
\end{equation}

\noindent where $T$ denotes the transpose, $\sqcup $ denotes the concatenate operation. $\sigma$ is the activation function, in our experiment, we use \emph{LeakyRelu} and the negative input slope ($s$ = 0.2) of the nonlinear activation function $\sigma$. To ensure the stability of the relationship between pairs of subjects, we extend the RAM to be multi-head, which can be expressed as:

\begin{equation}
\hat{\textbf{A}}_{i j}=\frac{1}{Q} \sum_{q=1}^{Q} \frac{\exp \left(\sigma\left(\xi_{q}^{T}\left[\psi _{q} \theta_{i} \sqcup  \psi _{q} \theta_{j}\right]\right)\right)}{\sum_{e \in N_{i}} \exp \left(\sigma\left(\xi_{q}^{T}\left[\psi _{q} \theta_{i} \sqcup  \psi _{q} \theta_{e}\right]\right)\right.},
\end{equation}

\noindent where $Q$ is the number of heads in the multi-head RAM, $\xi_{q}$ is the relational attention operator of the $q$-th head, and $\psi_{q}$ is the learnable weight parameter of the $q$-th head.

\section{EXPERIMENTAL RESULTS AND ANALYSIS}

\subsection{Experimental Settings}

\begin{table*}[!h]
\centering
\caption{The performance comparison of SOTA methods for ASD/TC classification.}
\label{tab1}
\renewcommand\arraystretch{1.4}
\begin{tabular}{lccccccr}
\toprule
\textbf{Method} & \textbf{ACC (\%)}        & \textbf{AUC (\%)}        & \textbf{SEN (\%)}        & \textbf{SPE (\%)}        & \textbf{Modal Type~}                         & \textbf{\textbf{Graph Type}}    & \textbf{\textbf{Time (sec)}}\\
\hline
PopGCN \cite{parisot2017spectral}        & 69.80 $\pm$ 3.35          &70.32 $\pm$ 3.90           & 73.35 $\pm$ 7.74          & 80.27 $\pm$ 6.48    & Single                                       & Static               & 31     \\
MultiGCN \cite{kazi2019self}        & 69.24 $\pm$ 5.90          & 70.04 $\pm$ 4.22          & 70.93 $\pm$ 4.68          & 74.33 $\pm$ 6.07          & Multiple                                     & Static              & 43     \\
InceptionGCN \cite{kazi2019inceptiongcn}  & 72.69 $\pm$ 2.37          & 72.81 $\pm$ 1.94          & 80.29 $\pm$ 5.10          & 74.41 $\pm$ 6.22          & Single                                       & Static        & 32           \\
LSTMGCN \cite{kazi2019graph}        & 74.92 $\pm$ 7.74          & 74.71 $\pm$ 7.92          & 78.57 $\pm$ 11.6          & 78.87 $\pm$ 7.79          & Multiple                                     & Static              & 39     \\
LG-GNN \cite{zhang2022classification}        & 81.75 $\pm$ 1.10          & 85.22 $\pm$ 1.01          & 83.22 $\pm$ 1.84          & 82.96 $\pm$ 0.94          & Multiple                                     & Static     & 134             \\
EVGCN \cite{huang2022disease}           & 85.90 $\pm$ 4.47          & 84.72 $\pm$ 4.27          & 88.23 $\pm$ 7.18          & 79.90 $\pm$ 7.37         & Multiple & Dynamic          & 61        \\
LGL \cite{cosmo2020latent}             & 86.40 $\pm$ 1.63          & 85.88 $\pm$ 1.75          & 86.31 $\pm$ 4.52          & 88.42 $\pm$ 3.04           & Multiple & Dynamic         & 56         \\
MMGL \cite{zheng2022multi}            & 89.77 $\pm$ 2.72          & 89.81 $\pm$ 2.56          & 90.32 $\pm$ 4.21          & 89.30 $\pm$ 6.04          & Multiple & Dynamic           & 67        \\
\hline
\textbf{MMKGL}   & \textbf{91.08 $\pm$ 0.59} & \textbf{91.01 $\pm$ 0.63} & \textbf{91.97 $\pm$ 0.64} & \textbf{90.05 $\pm$ 1.37}  & Multiple & Dynamic             & 112      \\
\bottomrule
\end{tabular}
\end{table*}

For a fair comparison, we performed a K-fold (K=5) cross-validation experiment on the ABIDE dataset. 
To be more specific, the dataset was split into 5 non-overlapping subsets. Each time we left one for test, and the rest for training and validation. In the training process, the model that performs best on the validation set was taken and evaluated on the test set. We repeated it 10 times, and the average performance was reported. 

Four widely used metrics were applied to evaluate the performance, including accuracy (ACC), area under the curve (AUC), sensitivity (SEN), and specificity (SPE).
In our experiment, hyper-parameters $\lambda_{1}$ and $\lambda_{2}$ were set to 1 empirically.


\subsection{Comparison with previous work}

We compare our approach with state-of-the-art (SOTA) methods, including PopGCN \cite{parisot2017spectral}, InceptionGCN \cite{kazi2019inceptiongcn}, MultiGCN \cite{kazi2019self}, LSTMGCN \cite{kazi2019graph}, LG-GNN \cite{zhang2022classification}, EVGCN \cite{huang2022disease}, LGL \cite{cosmo2020latent}, and MMGL \cite{zheng2022multi}. Among them, both PopGCN and InceptionGCN are early disease prediction studies based on single modal data and manual construction of static graphs. MultiGCN, LSTMGCN, LG-GNN, EVGCN, LGL, and MMGL are SOTA works that make use of multi-modal data for disease prediction. Moreover, EVGCN, LGL, and MMGL are dynamic in constructing graphs.

The experimental results are shown in Table~\ref{tab1}. 
As can be seen, approaches that use multi-modal generally outperform those use single modality. 
The average performance of the multi-modal approaches (82.72\%) is 11.48\% higher than the average performance of the single modal approach (71.24\%). Furthermore, the performance of the methods that use dynamic graphs, i.e., EVGCN, LGL, and MMGL, has a large performance improvement over the methods that use static graphs. 
This confirms the advantage of dynamic graphs over static graphs, i.e., sacrificing a negligible amount of training time in exchange for model performance as well as learnability.

Among the approaches that construct graphs dynamically, the MMGL method outperforms the EVGCN and LGL methods. One possible reason is that MMGL employs a cross-modal attention mechanism, which helps to capture valuable multi-modal complementary information. In contrast, our method outperforms MMGL in all four metrics. Compared to the traditional dynamic graph construction methods, our method uses multi-modal graph embedding to construct dynamic graphs, which alleviates the problem of negative effects between modal fusion. The supervision graph $\mathbf{S}_{SG}$ and function graph $\mathbf{S}_{FG}$ are also utilized to optimize the graph fusion process. In light of this, our dynamic graph method outperforms the SOTA methods.

We calculate the time spent on all methods to perform a single 5-fold cross-validation experiment. The results are shown in Table~\ref{tab1}. It is worth noting that the methods based on dynamic graph construction generally take longer than the static graph construction methods. This is because dynamic graph construction methods need to reconstruct the graph at each round of training. LG-GNN runs the longest time because it is based on graph classification, which requires separate graph construction for each sample. Compared to EVGCN, LGL and MMGL, MMKGL spends more time for a better performance, because it employs larger feature dimensions and a wider network structure resulting in a larger number of model parameters.



\subsection{Ablation Study}

\subsubsection{Effect of MMGE on MMKGL}

\begin{table*}[!h]
\centering
\caption{The performance of Modal Ablation of MMGE on MMKGL.}
\label{tab2}
\renewcommand\arraystretch{1.4}
\begin{tabular}{cllllccccc} 
\toprule
\multirow{2}{*}{\textbf{Method}}                                                           & \multicolumn{4}{c}{\textbf{Modal}}                                                                     & \multirow{2}{*}{\textbf{MKGL}} & \multirow{2}{*}{\textbf{ACC (\%)}} & \multirow{2}{*}{\textbf{AUC (\%)}} & \multirow{2}{*}{\textbf{SEN (\%)}} & \multirow{2}{*}{\textbf{SPE (\%)}}  \\ 
\cline{2-5}
                                                                                           & \multicolumn{1}{c}{PHE} & \multicolumn{1}{c}{Anat} & \multicolumn{1}{c}{Func} & \multicolumn{1}{c}{FC} &                                &                                    &                                    &                                    &                                     \\ 
\hline
\multirow{4}{*}{\begin{tabular}[c]{@{}c@{}}\textbf{Backbone}\\(Single Modal)\end{tabular}} & $\checkmark$            &                          &                          &                        &                                & 52.99 $\pm$ 0.96                   & 51.86 $\pm$ 0.97                   & 67.13 $\pm$ 2.35                   & 36.58 $\pm$ 2.62                    \\
                                                                                           &                         & $\checkmark$             &                          &                        &                                & 53.75 $\pm$ 1.61                   & 52.18 $\pm$ 1.64                   & 73.49 $\pm$ 3.41                   & 31.07 $\pm$ 4.11                    \\
                                                                                           &                         &                          & $\checkmark$             &                        &                                & 54.17 $\pm$ 0.64                   & 52.08 $\pm$ 0.74                   & 79.91 $\pm$ 3.80                   & 24.25 $\pm$ 4.51                    \\
                                                                                           &                         &                          &                          & $\checkmark$           &                                & 75.07 $\pm$ 0.87                   & 74.75 $\pm$ 0.88                   & 78.98 $\pm$ 1.47                   & 70.51 $\pm$ 1.78                    \\ 
\hline
\multirow{5}{*}{\begin{tabular}[c]{@{}c@{}}\textbf{MMKGL} \\(Multi-modal)\end{tabular}}    &                         &                          &                          & $\checkmark$           & $\checkmark$                   & ~77.76 $\pm$~2.13~                 & ~77.55 $\pm$~2.15~                 & 81.15 $\pm$ 1.60                   & 73.95 $\pm$ 3.31                    \\
                                                                                           & $\checkmark$            &                          &                          & $\checkmark$           & $\checkmark$                   & 81.81 $\pm$ 1.52                   & 81.41 $\pm$ 1.49                   & 86.30 $\pm$ 2.47                   & 76.52 $\pm$ 2.38                    \\
                                                                                           & $\checkmark$            &                          & $\checkmark$             & $\checkmark$           & $\checkmark$                   & 87.87 $\pm$ 0.63                   & 87.58 $\pm$ 0.72                   & 90.67 $\pm$ 0.91                   & 84.48 $\pm$ 2.06                    \\
                                                                                           & $\checkmark$            & $\checkmark$             &                          & $\checkmark$           & $\checkmark$                   & 88.47 $\pm$ 0.96                   & 88.36 $\pm$ 0.93                   & 90.86 $\pm$ 1.91                   & 85.85 $\pm$ 1.28                    \\
                                                                                           & $\checkmark$            & $\checkmark$             & $\checkmark$             & $\checkmark$           & $\checkmark$                   & \textbf{91.08 $\pm$ 0.59}          & \textbf{91.01 $\pm$ 0.63}          & \textbf{91.97 $\pm$ 0.64}          & \textbf{90.05 $\pm$ 1.37}           \\
\bottomrule
\end{tabular}
\end{table*}

To evaluate the effectiveness of different modalities, we separately input four modalities FC, Anat, Func, and PHE into the two-layer graph convolutional network (backbone). As shown in Table \ref{tab2}, all four modalities demonstrate their effectiveness for autism prediction, with FC exhibiting the strongest representation ability and achieving an accuracy of 75.07\%, the highest among the four modalities.

Next, we evaluate the performance of modal combinations based on MMKGL. We replace the backbone with MKGL and add other modalities incrementally, with FC as the primary modality. As shown in Table \ref{tab2}, the accuracy of MKGL+FC is 77.76\%, which represents a 2.69\% improvement over the backbone's accuracy (75.07\%). Considering that Anat and Func are imaging information, while PHE belongs to clinical phenotype information, we first incorporate PHE into MKGL+FC, resulting in a performance of 81.81\%. This indicates that PHE can effectively complement FC and improve the prediction performance of the model. We subsequently add Anat and Func to MKGL+FC+PHE, and the performance of MKGL+FC+PHE+Anat and MKGL+FC+PHE+Func reaches 88.47\% and 87.87\%, respectively. This suggests that Anat and Func can synergize well with the FC+PHE modality. The four modalities (FC+Func+Anat+PHE) demonstrate a higher accuracy of 91.08\% than the three modalities, indicating that FC, Func, Anat, and PHE can be effectively integrated through MMGE.

\subsubsection{Effect of MKGL on MMKGL}

To validate the effectiveness of Multi-Kernel Graph Learning (MKGL) module, we fix the MMGE module and investigate the contribution of different components of MKGL.
\begin{enumerate}[(1)]
    \item GCN: It uses the graph convolution network only.
    \item GCN + CKDT: It adds the CKDT to GCN.
    \item GCN + RAM: It integrates the RAM to GCN.
    \item MKGL: it combines both RAM and CKDT with GCN, i.e., GCN+RAM+CKDT.
\end{enumerate}

\begin{table*}[!h]
\centering
\caption{The performance of Module Ablation of MKGL on MMKGL.}
\label{tab3}
\renewcommand\arraystretch{1.4}
\begin{tabular}{ccccccccc} 
\toprule
\multirow{2}{*}{\textbf{Method}} & \multirow{2}{*}{~\textbf{ MMGE}~~} & \multicolumn{3}{c}{\textbf{MKGL}}         & \multirow{2}{*}{\textbf{ACC (\%)}} & \multirow{2}{*}{\textbf{AUC (\%)}} & \multirow{2}{*}{\textbf{SEN (\%)}} & \multirow{2}{*}{\textbf{SPE (\%)}}  \\ 
\cline{3-5}
                                 &                                    & GCN &~ RAM~~      & \textbf{~ }CKDT\textbf{~}~ &                                    &                                    &                                    &                                     \\ 
\hline
\multirow{4}{*}{\textbf{MMKGL}}  & $\checkmark$  &      $\checkmark$               &              &                            & 80.61 $\pm$ 1.24                   & 80.66 $\pm$ 1.23                   & 79.94 $\pm$ 2.12                   & 81.39 $\pm$ 1.99                    \\
                                 & $\checkmark$       &$\checkmark$                &              & $\checkmark$               & 81.13 $\pm$ 1.12                   & 81.15 $\pm$ 1.17                   & 80.79 $\pm$ 1.90                   & 81.51 $\pm$ 2.73                    \\
                                 & $\checkmark$       &$\checkmark$                & $\checkmark$ &                            & 89.45 $\pm$ 0.79                   & 89.37 $\pm$ 0.78                   & 90.47 $\pm$ 1.32                   & 88.26 $\pm$ 1.16                    \\
                                 & $\checkmark$                       & $\checkmark$ & $\checkmark$ &$\checkmark$              & \textbf{91.08 $\pm$ 0.59}          & \textbf{91.01 $\pm$ 0.63}          & \textbf{91.97 $\pm$ 0.64}          & \textbf{90.05 $\pm$ 1.37}           \\
\bottomrule
\end{tabular}
\end{table*}

The experimental results are shown in Table \ref{tab3}. It can be observed that the accuracy of using GCN alone is 80.61\%. By separately adding CKDT and RAM to GCN, the accuracy of GCN+CKDT and GCN+RAM improve to 81.13\% and 89.45\%, respectively. It indicates that RAM and CKDT effectively utilize the multi-modal graph $\mathcal{G}$ generated by MMGE to extract discriminative information for autism prediction. The accuracy of GCN+RAM+CKDT (MKGL) is 91.08\%, which validates the effectiveness of MKGL. Using MKGL, our model is improved by 10.47\%, which shows the promising ability to combine GCN, RAM, and CKDT.

\subsubsection{Effect of Feature Fusion Method}
To verify the effectiveness of the feature fusion method in CKDT, we replaced it with feature concatenation (Concat), feature addition (Add), feature weighting (Weight), and feature averaging (Avg). For feature weighting, we obtain the optimal expression by assigning different weights to the multi-modal features. 

\begin{figure}[!h]

\centerline{\includegraphics[width=\columnwidth]{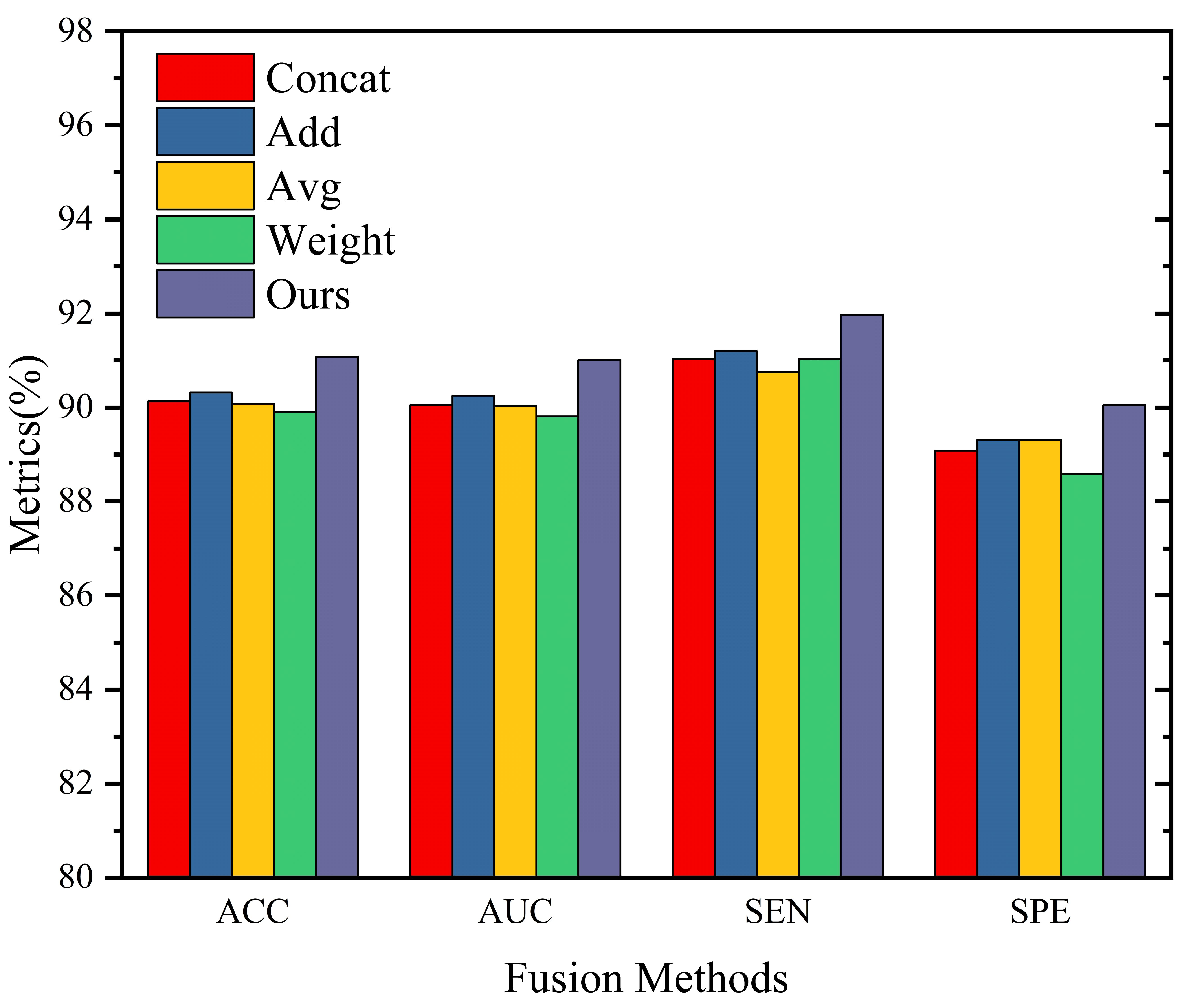}}
\caption{Performance comparison of feature fusion method in multi-kernel graph learning and traditional methods.}
\label{fig2}
\end{figure}

The experimental results are shown in Fig.~\ref{fig2}. In terms of accuracy ranking, the performance of the fusion methods from high to low is Ours, Add, Concat, Avg, and Weight. Unexpectedly, the performance of the feature weighting fusion method is the worst, even lower than that of Avg. This may be due to the model learning inappropriate modality weights, which leads to relatively poor performance. According to previous research, the feature weighting fusion method generally performs well in these traditional feature fusion methods. It is worth noting that our fusion module outperforms traditional methods on all evaluation metrics. This indicates that the fusion method in CKDT can effectively integrate heterogeneous information on the graph.

\subsection{Hyperparameters Analysis}
\subsubsection{Multiple Convolutional Kernel Combinations Analysis}

To investigate the influence of convolutional kernel size on the model, we evaluate the performance of single convolutional kernels separately using a single graph convolutional network (MMGE+RAM). 
$K$ is the convolution kernel receptive field size, in our experiment, $K = \left \{  x\ |\ x\in N  ,\ 1 \le x \le 5\right\}$. Subsequently, we test a combination (e.g., 1+2+3, 2+3+4,...) of three randomly selected convolutional kernels from the $K$ convolutional kernels. 
\begin{figure}[!t]
\centerline{\includegraphics[width=\columnwidth]{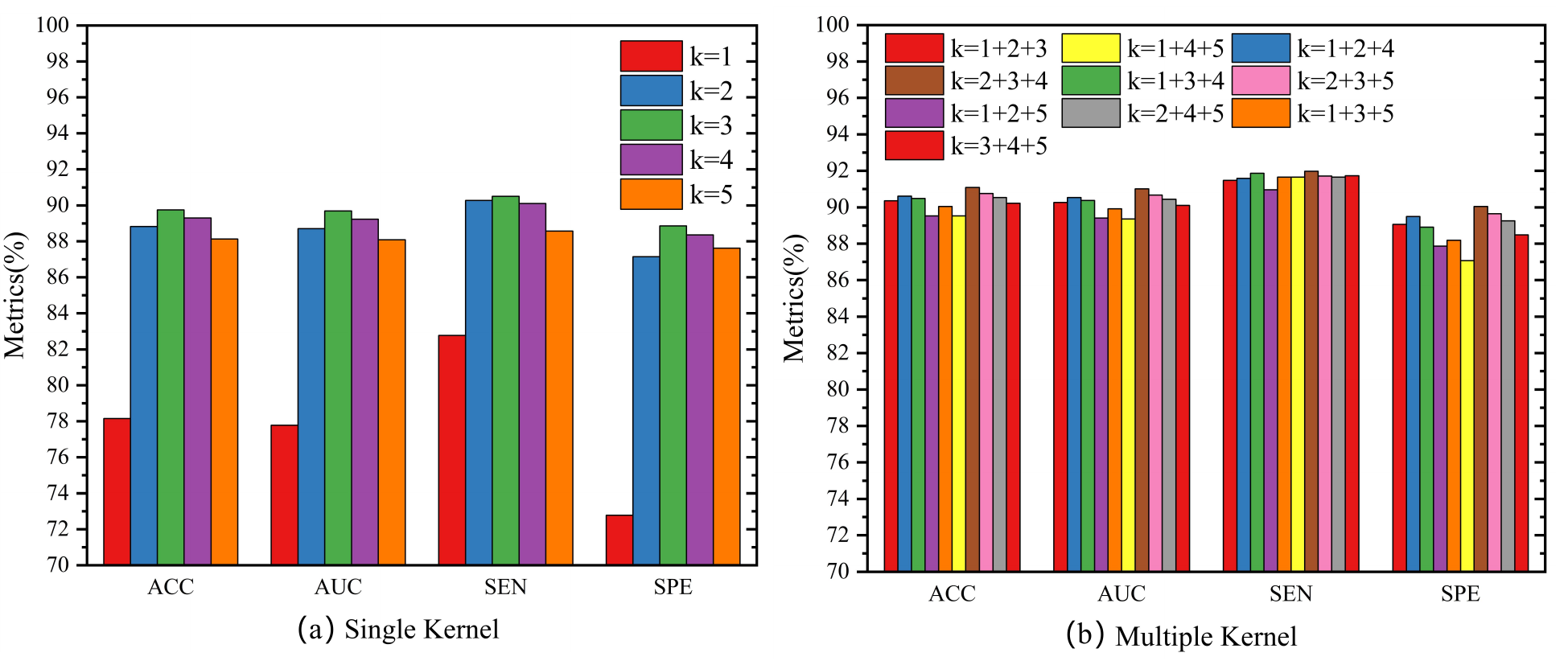}}
\caption{(a) Performance comparison of single convolutional kernel receptive field size. (b) Performance comparison of multiple combinations of convolutional kernels with different receptive field sizes.}
\label{cheb}
\end{figure}

The experimental results are shown in Fig.~\ref{cheb}(a).
As can be seen from, the performance of the model shows a trend of increasing and then decreasing with the increase of the receptive field $K$. The model performance reaches the maximum when $K$=3. 
Too small convolutional kernel size, i.e., $K$=1, cannot effectively extract information from the multi-modal graph $\mathcal{G}$, and too large convolutional kernels, e.g., $K$=5, causes an oversmoothing effect. As can be seen from Fig.~\ref{cheb}(b), model achieves the best performance among all combinations when the convolutional kernel combination is $K$=2+3+4. This also confirms the conclusion obtained in Fig. \ref{cheb}(a), i.e., the closer the convolution kernel size $K$ is to 3, the better the performance.

\subsubsection{Training Set Ratio Analysis}
It is well known that one challenge in the field of deep learning in medicine is the lack of training data. In this context, we explore the performance of the model with a smaller dataset. In the normal case, we usually use the traditional data partitioning approach, i.e., we divide the dataset into training, validation, and test sets in the ratio of 60\%, 20\%, and 20\%, respectively. 

\begin{figure}[!h]
\centerline{\includegraphics[width=\columnwidth]{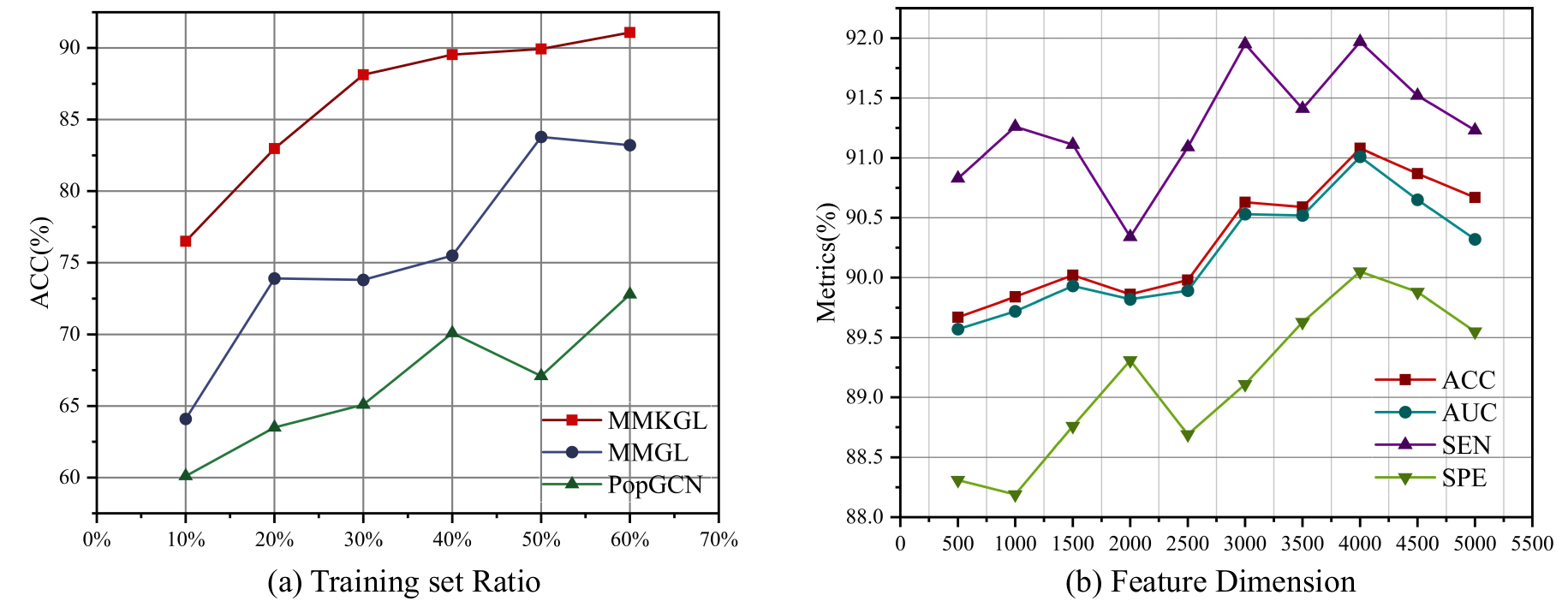}}
\caption{(a) Accuracy comparison of multiple methods with different training set ratio. (b) Performance Comparison with different feature input dimensions of MMKGL.}
\label{fig3}
\end{figure}

As shown in Fig.~\ref{fig3}(a), we set the training set ratio from 10\% to 60\%, and compare our method with MMGL and PopGCN with the same training set ratio, it is obvious that our method performs much better than the other two methods in the same ratio. Compared with MMGL and PopGCN that fluctuate largely when training set ratio increases from 30\% to 60\%, our method shows a smooth and stable increase. It is worth mentioning that in the lowest 10\% percentage, our method performs much better than MMGL and PopGCN, which indicates that our method is well suited for the domain with small data volume.

\subsubsection{Feature Dimension Analysis}
Feature dimensions and model performance are closely related. To analyze the impact of different feature dimensions on model performance and determine the optimal feature dimensions, the feature dimension is increased from 500 to 5000 with a step of 500. The model performance with different feature dimensions is reported.

The experimental results are shown in Fig.~\ref{fig3}(b). In general, the performance of the model shows a gradual increase but with a fluctuation when the feature dimension is 2000, where the SPE performance starts to decrease while the SEN starts to increase. This may be due to the imbalance of the proportion of positive and negative samples in the data. The optimal performance is reached when the feature dimension is equal to 4000.

\section{Discussions}
\begin{figure}[h]   
	\centering
	
	\includegraphics[width=\linewidth,scale=1.00]{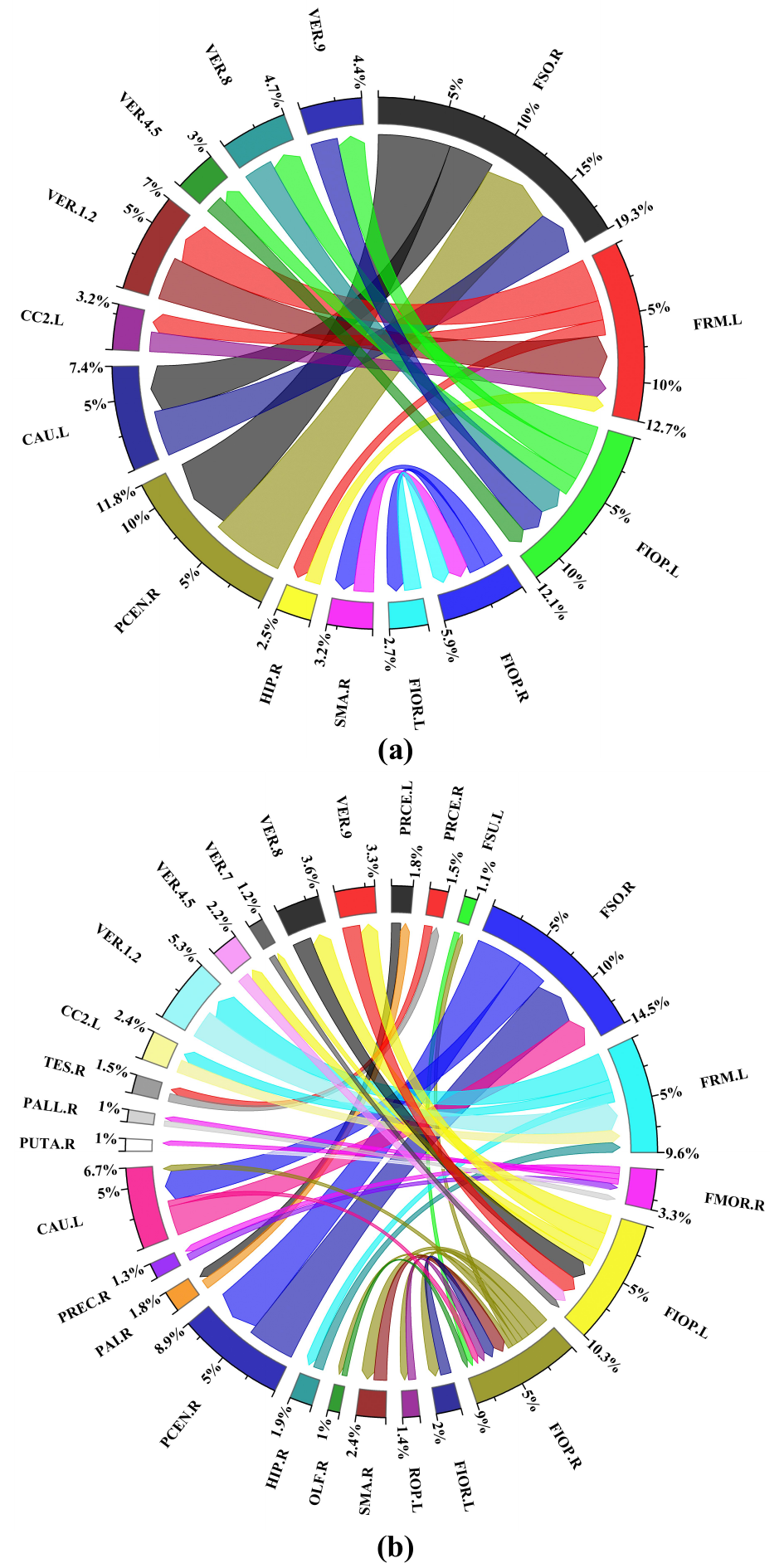}
	\caption{The figure shows the most discriminative functional connections extracted from the model weights by the gradient-based approach. The thickness of the connections represent their weights. (a) Top 10 discriminative brain functional connectivities. (b)  Top 20 discriminative brain functional connectivities.}
	
	\label{fc}
\end{figure}

In this section, we analyze the brain regions and their constituent subnetworks that are of significant discriminatory for autism diagnosis in our model. Specifically, we use a gradient-based approach \cite{selvaraju2017grad} to present FC self-attentive weighting coefficients from the model for all subjects and to obtain the average attentional level of these FCs. Notably, we used the automatic anatomical labeling (AAL) atlas and select the top K functional connections that are most discriminatory, respectively. The top 10 and top 20 most discriminating functional connections, standardized by their weight scores, are shown in Fig. \ref{fc}. 

\subsection{Discriminative Brain Regions}

As shown in Table \ref{tab4}, the discriminative brain areas that distinguish autism from healthy controls included the following: Frontal Area) FSO.R, FRM.L, FIOP.R, FMOR.R. Motorium) CAU.L. Sensorium) PCEN.R. Cerebellum) VER.1.2, VER.4.5, VER.8, VER.9, CC2.L. There are also some important brain areas that have a relatively small weighting, but still contribute to the diagnosis of autism, such as FIOR.L, SMA.R, HIP.R, PAI.R, PRCE.L, PRCE.R. Below we discuss the role of the discriminative brain regions that we find in the pathogenesis pattern of autism of previous studies.

\begin{figure*}
	
	\centering
	
	\includegraphics[width=\linewidth,scale=1.00]{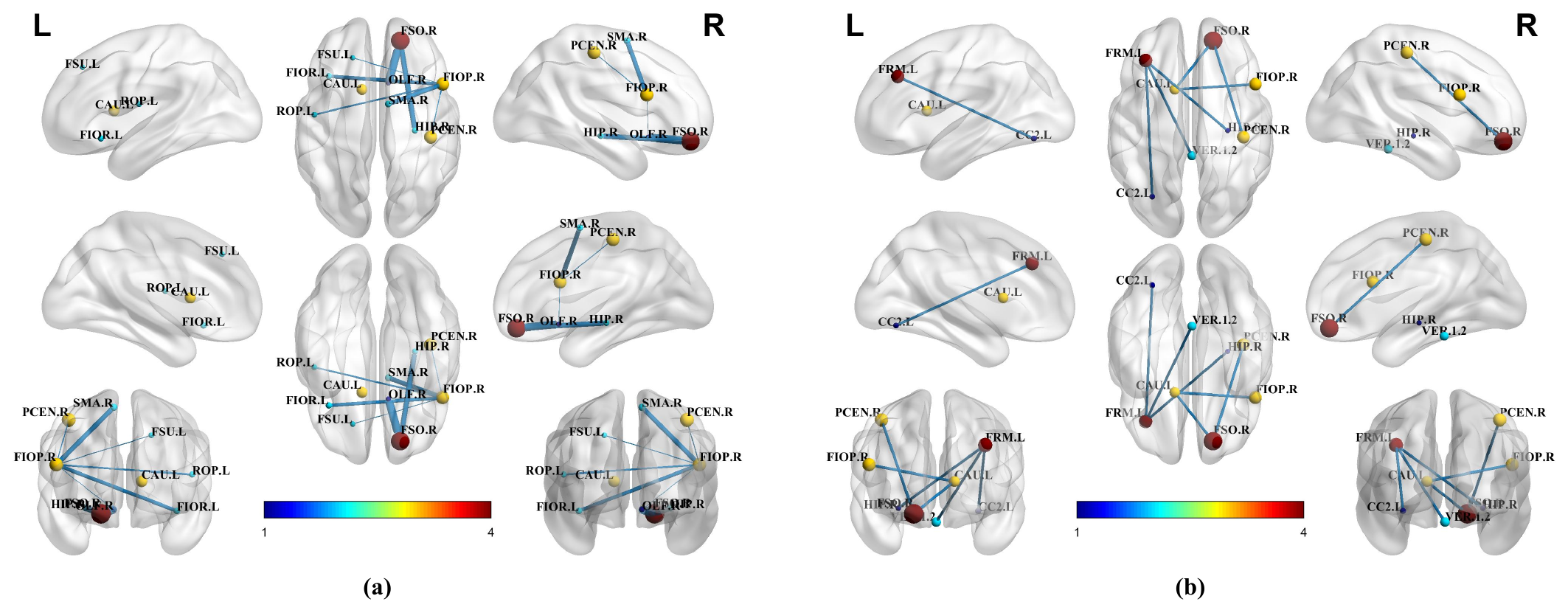}
	\caption{Visualization of the brain network composed of the top discriminative brain functional connectivities. Each figure contains views from different layouts (sagittal, axial, coronal). The color bars represent the weights of the brain regions and the thickness of the connections represent their weights. (a) Discriminative subnetworks with FIOP.R as the center of functional connectivity. (b) Discriminative subnetworks with FSO.R and FRM.L as the center of the functional connectivity.}
	\label{subnet}
	
\end{figure*}

In a study of related literature, it was found that in the frontal orbit (FSO.R, FIOR.L), patients with asd respond to mildly aversive visual and auditory stimuli \cite{green2015neurobiology}. There are abnormalities in the morphological structure of the ossicle (FIOP) in patients with autism and normal people, and there is a correlation with social barriers. ASD's palpebral activity is relatively calm \cite{yamasaki2010reduced}.
Frontal middle gyrus (FRM.L) gene expression in ASD is different from normal \cite{crider2014dysregulation}. The volume of the caudate nucleus is enlarged in medication-naive subjects with autism. The strength of connectivity within and between different functional subregions of the precentral gyrus (PRCE) is associated with the diagnosis of ASD and the severity of ASD features \cite{samson2012enhanced}. The postcentral gyrus (PCEN) is responsible for somatosensory sensation. The postcentral gyrus cortical thickness and gray matter concentration are reduced in autistic subjects, and research has determined that the postcentral gyrus is a key brain area for ASD \cite{fatemi2018metabotropic}. Cerebellar worms have been implicated in the regulation of limbic functions, including mood, sensory reactivity, and salience detection. Association study between posterior earthworm and mesocerebellar cortex suggests cerebellum plays key role in autism \cite{fatemi2012consensus}.

\begin{table}[!h]
\centering
\caption{The ROIS selected for the most discriminative functional connection and its normalized weight score.}
\label{tab4}
\renewcommand\arraystretch{1.3}
\resizebox{\columnwidth}{!}{
\begin{tabular}{ccccc}
\toprule
\textbf{ID} & \textbf{ROI} & \textbf{Anatomical Region} & \multicolumn{1}{l}{\textbf{FC-10 Weight}} & \multicolumn{1}{l}{\textbf{FC-20 Weight}}  \\
\hline
1           & PRCE.L       & Precentral\_L              & -                                         & 1.76\%                                     \\
2           & PRCE.R       & Precentral\_R              & -                                         & 1.53\%                                     \\
3           & FSU.L        & Frontal\_Sup\_L            & -                                         & 1.06\%                                     \\
6           & FSO.R        & Frontal\_Sup\_Orb\_R       & 19.26\%                                   & 14.51\%                                    \\
7           & FRM.L        & Frontal\_Mid\_L            & 12.72\%                                   & 9.58\%                                     \\
11          & FIOP.L       & Frontal\_Inf\_Oper\_L      & 12.09\%                                   & 10.32\%                                    \\
12          & FIOP.R       & Frontal\_Inf\_Oper\_R      & 5.93\%                                    & 8.99\%                                     \\
15          & FIOR.L       & Frontal\_Inf\_Orb\_L       & 2.71\%                                    & 2.04\%                                     \\
17          & ROP.L        & Rolandic\_Oper\_L          & -                                         & 1.40\%                                      \\
20          & SMA.R        & Supp\_Motor\_Area\_R       & 3.22\%                                    & 2.43\%                                     \\
22          & OLF.R        & Olfactory\_R               & -                                         & 0.96\%                                     \\
26          & FMOR.R       & Frontal\_Mid\_Orb\_R       & -                                         & 3.31\%                                     \\
38          & HIP.R        & Hippocampus\_R             & 2.49\%                                    & 1.87\%                                     \\
58          & PCEN.R       & Postcentral\_R             & 11.84\%                                   & 8.92\%                                     \\
62          & PAI.R        & Parietal\_Inf\_R           & -                                         & 1.76\%                                     \\
68          & PREC.R       & Precuneus\_R               & -                                         & 1.28\%                                     \\
71          & CAU.L        & Caudate\_L                 & 7.42\%                                    & 6.69\%                                     \\
74          & PUTA.R       & Putamen\_R                 & -                                         & 1.02\%                                     \\
76          & PALL.R       & Pallidum\_R                & -                                         & 1.01\%                                     \\
82          & TES.R        & Temporal\_Sup\_R           & -                                         & 1.53\%                                     \\
93          & CC2.L        & Cerebelum\_Crus2\_L        & 3.20\%                                    & 2.41\%                                     \\
109         & VER.1.2      & Vermis\_1\_2               & 7.03\%                                    & 5.30\%                                      \\
111         & VER.4.5      & Vermis\_4\_5               & 2.98\%                                    & 2.25\%                                     \\
113         & VER.7        & Vermis\_7                  & -                                         & 1.21\%                                     \\
114         & VER.8        & Vermis\_8                  & 4.74\%                                    & 3.57\%                                     \\
115         & VER.9        & Vermis\_9                  & 4.37\%                                    & 3.29\%                                     \\
\bottomrule
\end{tabular}}
\end{table}

\subsection{Discriminative subnetworks}

We construct 2 subnetworks, as shown in Fig.~\ref{subnet}(b). They are the subnetwork FPCF and FCVH. FPCF connects PCEN.R, CAU.L, and FIOP.R with the brain area FSO.R as the core. According to the research of \cite{green2015neurobiology}, the blood oxygen signals of the amygdala and the frontal eye (FSO.R) in adolescents with ASD showed significant correlation changes under sensory stimulation. The amygdala and caudate nucleus (CAU.L) are directly connected in the brain. In the subnetwork we found in FPCF, CAU.L and FSO.R are connected. This may suggest that CAU.L provides a bridge for the connection between the amygdala and FSO.R. According to \cite{glerean2016reorganization}, CAU.L is clearly connected to PCEN.R in the Ventro-temporal limbic (VTL) subnetwork that differs most between the ASD and normal groups. At the same time, in VTL, CAU.L also has a certain connection with the amygdala.

The subnetwork FCVH connects CC2.L, VER.1.2, and HIP.R with the brain region FRM.L as the core. Numerous studies from neurocognitive and neuroimaging have shown that FRM is associated with the pathophysiology of ASD \cite{barendse2013working}. In addition, there is bisexual dimorphism in the middle frontal gyrus \cite{goldstein2001normal}, which may be why men are four times more likely to develop autism than women. According to research by \cite{fatemi2012consensus}, neuropathological abnormalities in autism were found in the cerebellum (CC2.L). In the subnetwork FCVH, both FRM.L and HIP.R are related top memory, and FRM.L is mainly responsible for short-term memory.

According to Table \ref{tab4}, there is an abnormal phenomenon, that is, the weight of the brain region FIOP.R has increased by 3.06\%. This shows that on the basis of FC-10, most of the newly added functional connections in FC-20 are related to FIOP.R. In this regard, we use FIOP.R as the core to visualize the brain regions (FSU.L, ROP.L, SMA.R, OLR.R, HIP.R, CAU.L, FSO.R) connected to it, The results are shown in Fig. \ref{subnet}(a). According to the study by \cite{yamasaki2010reduced}, impaired social skills in autistic patients are associated with reduced gray matter volume at the FIOP site. ROP.L has also been shown to be associated with autism in the subnetwork AUD discovered by \cite{glerean2016reorganization}. \cite{carper2005localized} found that the FSU.L region of autism patients was significantly enlarged compared with controls. \cite{enticott2009electrophysiological}. identified SMA.R as a possible source of motor dysfunction in autism by examining motor-related potentials (MRPs).

\begin{figure}[h]
\centerline{\includegraphics[width=\columnwidth]{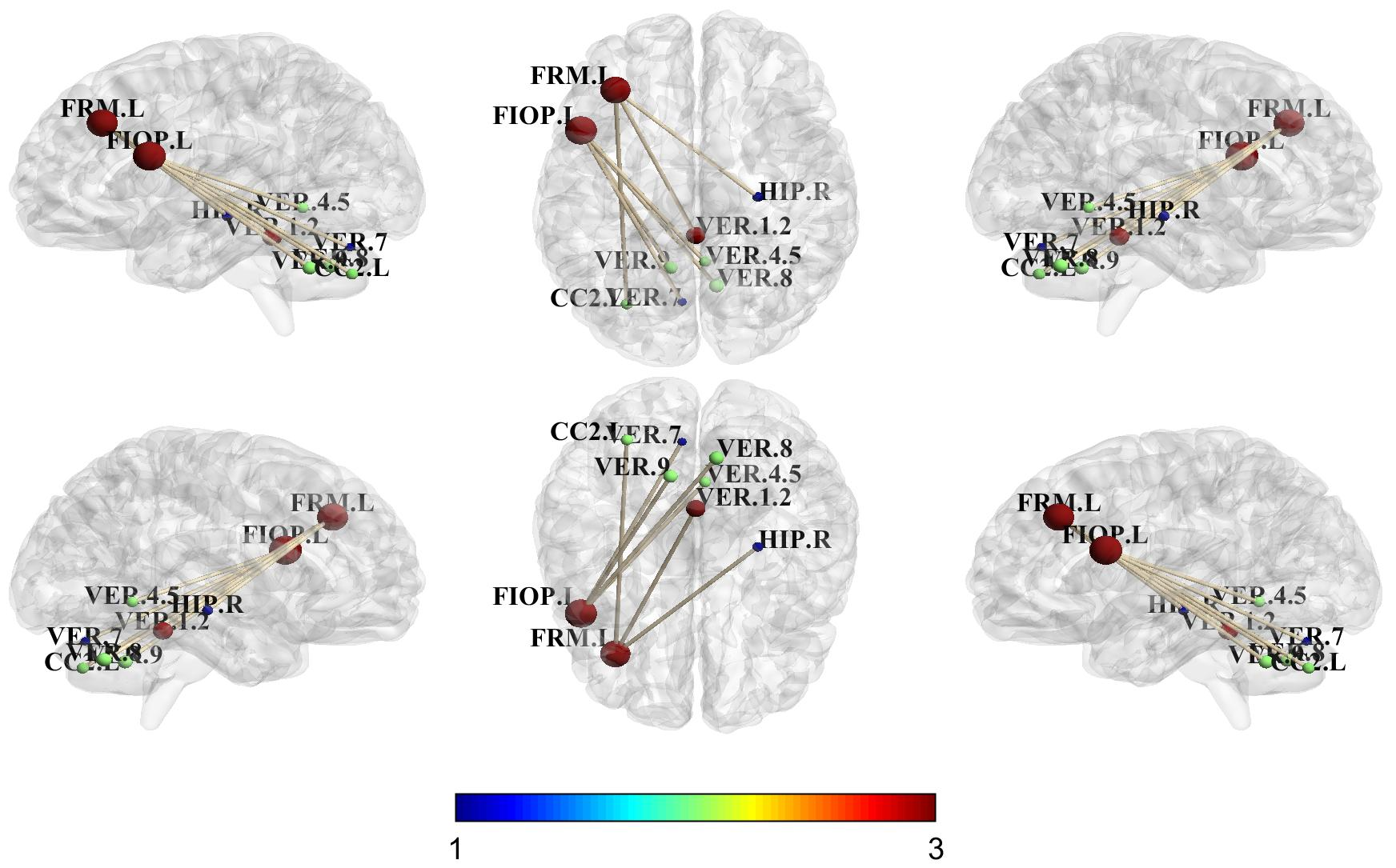}}
\caption{ Discriminative subnetwork of the cerebellum and the vermis.}
\label{Vermis}
\end{figure}

As shown in Fig. \ref{fc}(a), the brain regions related to the cerebellum account for a larger proportion. We visualize the brain regions of the cerebellum (CC2.L, VER.1.2, VER.4.5, VER.7, VER.8, VER.9) and their connected FIOP.L, FRM.L. The results are as follows shown in Fig. \ref{Vermis}. As the central part of the cerebellum, vermis (VER) has the functions of regulating muscle tone, maintaining the balance of the body, and coordinating movements. \cite{stanfield2008towards} concluded that the area of lobules I-V and VI-VII of the vermis are reduced in individuals with autism compared to controls. 
Loss of Purkinje cells in the posterior vermis and cerebellar intermediate cortex is the most consistent neuropathology in post-mortem dissection studies of the brains of individuals with autism \cite{webb2009cerebellar}. For more information on the pathological mechanism of the cerebellum in autism, please refer to \cite{fatemi2012consensus}.

Static network biomarkers have limitations in recognizing predisease samples and therefore lack the ability for early diagnosis. Dynamic network biomarkers (DNB) based on nonlinear dynamics can well distinguish the developmental state of diseases and realize the great potential of early diagnosis of diseases. To obtain early warning signals of the disease, Chen et al.~\cite{chen2012detecting} theoretically derived a DNB-based index, which indicates a sudden deterioration before a shift in the critical state of the disease. To prevent further irreversible deterioration of hepatocellular carcinoma, Yang et al.~\cite{yang2018dynamic} developed predictive dynamic network biomarkers capable of detecting the critical point. In the future, we will further explore dynamic network biomarkers regarding autism prediction.

\section{Conclusion}

In this study, we propose multi-modal multi-kernel graph learning for autism prediction and biomarker discovery. Our proposed multi-modal graph embedding is well suited to alleviate the problem of negative effects between modalities. In addition, our proposed multi-kernel graph learning network is capable of extracting heterogeneous information from multi-modal graphs for autism prediction and biomarker discovery. Finally, we find some important brain regions and subnetworks with important discriminatory properties for autism by a gradient-based approach. These findings provide important guidance for the study of autism pathological mechanisms.

\section{ACKNOWLEDGMENT}
This work was supported in part by the National Natural Science Foundation of China under Grants 62172444 and U22A2041, the Shenzhen Science and Technology Program under Grant KQTD20200820113106007, the Natural Science Foundation of Hunan Province under Grant 2022JJ30753, the Xinjiang Key Research and Development project 2023B01032, the Central South University Innovation-Driven Research Programme under Grant 2023CXQD018, and the High Performance Computing Center of Central South University.

\bibliographystyle{IEEEtran}
\bibliography{refs}

\begin{IEEEbiography}[{\includegraphics[width=1in,height=1.25in,clip,keepaspectratio]{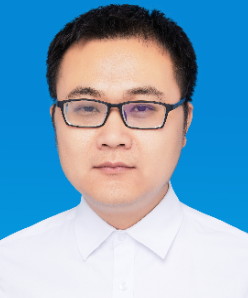}}]{Jin Liu}
 (Member, IEEE) received the PhD degree in computer science and technolgy from Central South University, China, in 2017. He is currently a professor at the School of Computer Science and Engineering, Central South University, Changsha, China. His current research interests include medical image computing, deep learning, and multi-modal learning. His research has been cited over 3500 times (Google Scholar) with H-index=28.
 
\end{IEEEbiography}

\begin{IEEEbiography}[{\includegraphics[width=1in,height=1.25in,clip,keepaspectratio]{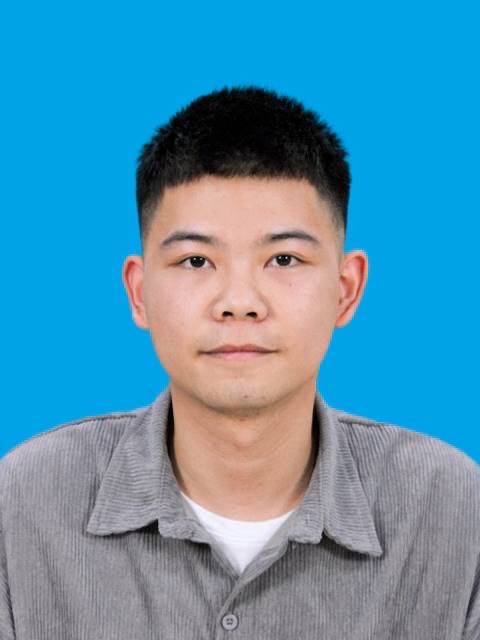}}]{Junbin Mao} received the B.S. degree from Hunan Agricultural University, Changsha, China, in 2021, and the M.S. degree from Central South University, Changsha, China. He is currently pursuing the Ph.D. degree with the School of Computer Science and Engineering, Central South University. His research interests include medical image analysis and deep learning.
\end{IEEEbiography}

\begin{IEEEbiography}[{\includegraphics[width=1in,height=1.25in,clip,keepaspectratio]{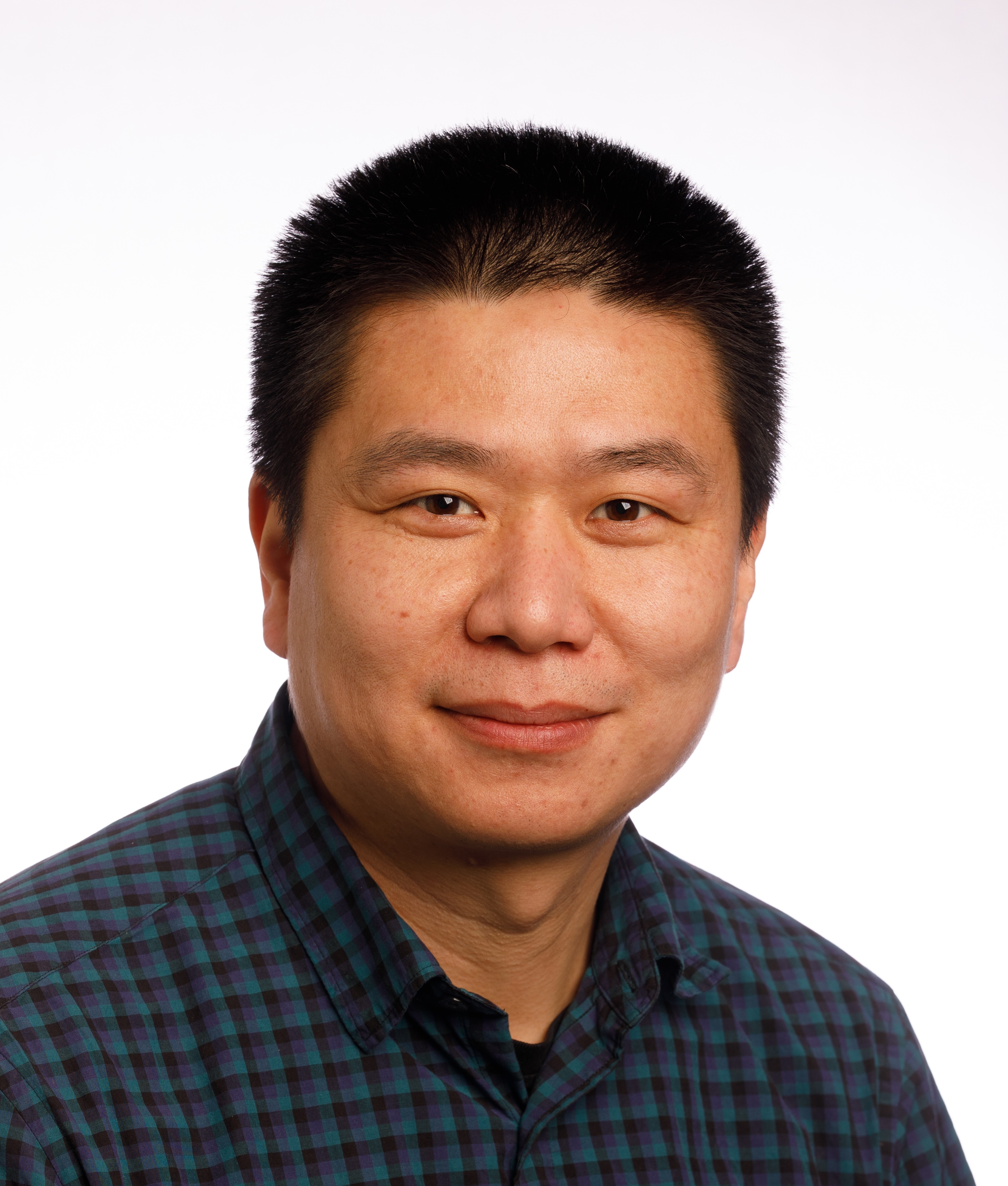}}]{Hanhe Lin} (Senior Member, IEEE) 
obtained his PhD degree from the University of Otago in 2016. From 2016 to 2021, he was a postdoctoral researcher at the University of Konstanz, funded by the German Research Foundation (DFG). After being a research fellow at the National Subsea Centre (Robert Gordon University) for a short period, he is currently a Lecturer in Computing at the School of Science and Engineering, University of Dundee. As an active researcher in research of computer vision, image processing, and visual quality assessment, he has published over 50 peer-reviewed papers. He also served as a technical program committee member or a reviewer in numerous prestigious conferences (e.g., QoMEX, ICIP, and ICME) and journals (e.g., IEEE Trans. Pattern Analysis and Machine Intelligence, IEEE Trans. Image Processing, and IEEE Trans. Multimedia.)
\end{IEEEbiography}

\begin{IEEEbiography}[{\includegraphics[width=1in,height=1.25in,clip,keepaspectratio]{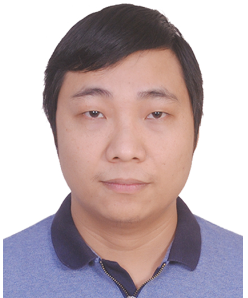}}]{Hulin Kuang}
 (Member, IEEE) received the B.Eng. and M.Eng. degrees from Wuhan University, Wuhan, China, in 2011 and 2013, respectively, and the Ph.D. degree from the City University of Hong Kong,
Hong Kong, in 2016.

He was a Post-Doctoral Fellow with the Department of Clinical Neurosciences, University of Calgary, Calgary, AB, Canada. He is currently an Associate Professor at the School of Computer, Central South University, Changsha, China. His research interests include deep learning, intelligent transportation systems, and medical image processing.
\end{IEEEbiography}

\begin{IEEEbiography}[{\includegraphics[width=1in,height=1.25in,clip,keepaspectratio]{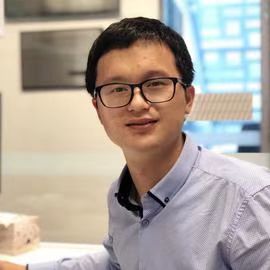}}]{Shirui Pan}
received the Ph.D degree in computer science from the University of Technology Sydney (UTS), Ultimo, NSW, Australia. He was a lecturer with the School of Software, UTS. He is currently a senior lecturer with the Faculty of Information Technology, Monash University, Clayton, VIC, Australia. He has authored or coauthored more than 60 research articles in top-tier journals and conferences, including the IEEE Transactions on Neural Networks and Learning Systems, the IEEE Transactions on Knowledge and Data Engineering, the IEEE Transactions on Cybernetics, the IEEE International Conference on Data Engineering, the AAAI Conference on Artificial Intelligence, the International Joint Conferences on Artificial Intelligence, and the IEEE International Conference on Data Mining . His research interests include data mining and machine learning.
\end{IEEEbiography}

\begin{IEEEbiography}
[{\includegraphics[width=1in,height=1.25in,clip,keepaspectratio]{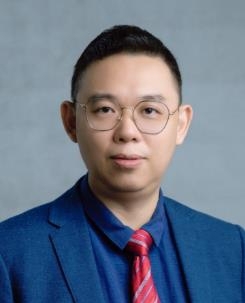}}]{Xusheng Wu}
received the Ph.D.degree in Busi- ness Administration from the University of Chica- go, Illinois, USA, in 2022. 

He is currently the Deputy Director of the Shenzhen Health Development Research and Data Management Center. His research interests include medical artificial intelligence research, health big data analysis, and intelligent medical engineering applications.
\end{IEEEbiography}

\begin{IEEEbiography}
[{\includegraphics[width=1in,height=1.25in,clip,keepaspectratio]{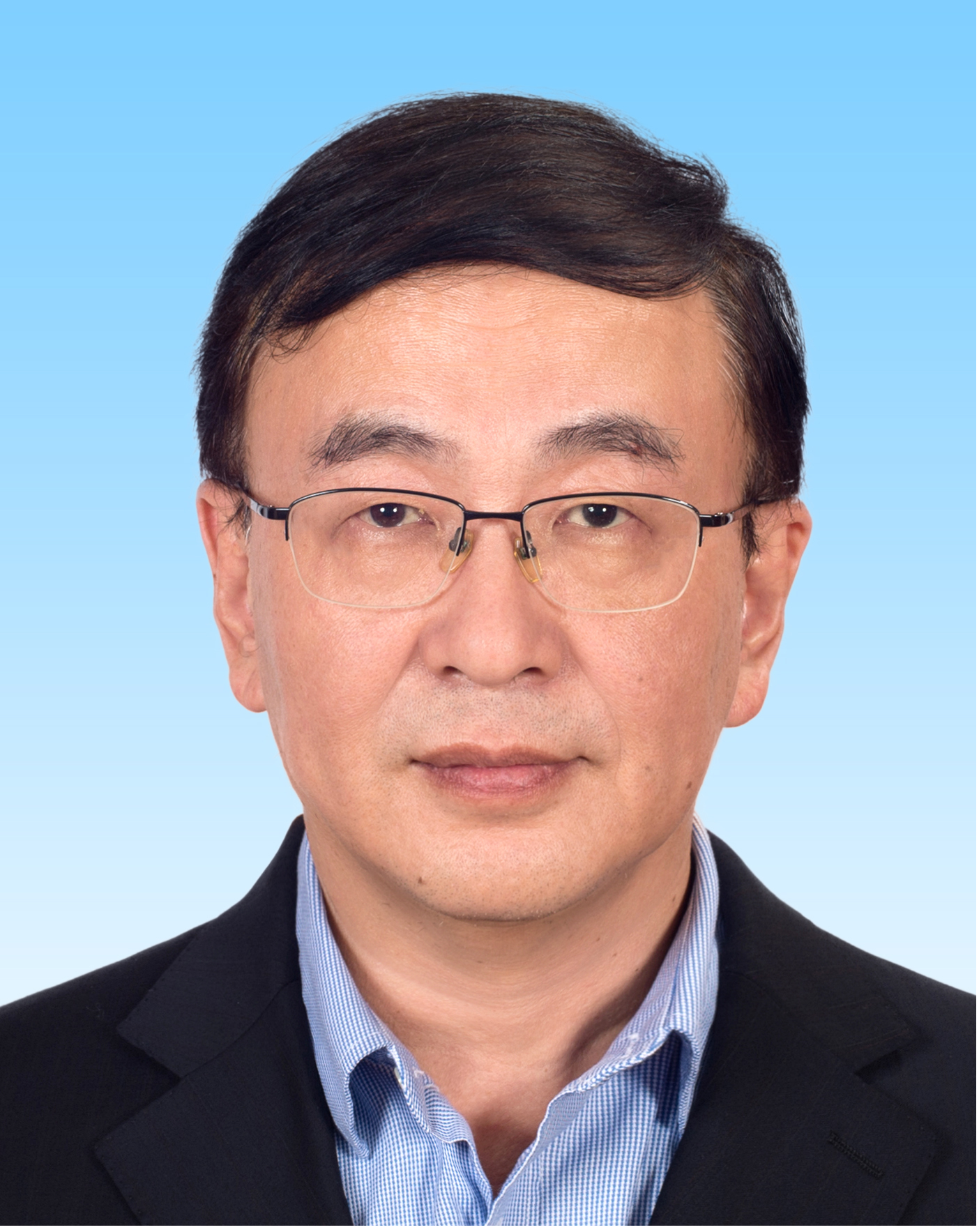}}]{Shan Xie}
received the Ph.D.degree in Busi- ness Administration from the University of Chica- go, Illinois, USA, in 2022. 

He is currently the Deputy Director of the Shenzhen Health Development Research and Data Management Center. His research interests include medical artificial intelligence research, health big data analysis, and intelligent medical engineering applications.
\end{IEEEbiography}

\begin{IEEEbiography}
[{\includegraphics[width=1in,height=1.25in,clip,keepaspectratio]{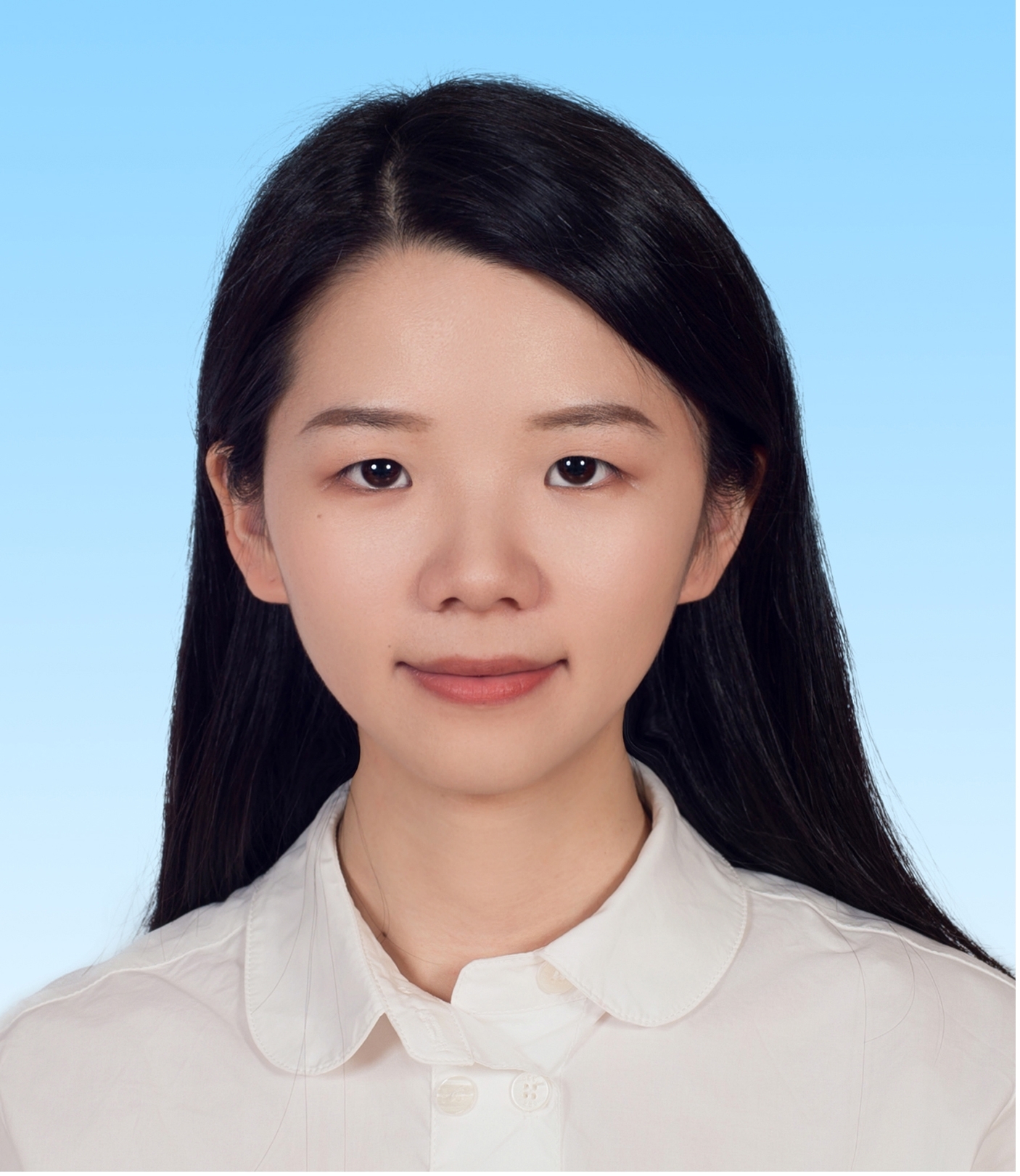}}]{Fei Liu }
received the Master's degree in Medical Information Management from Central South University, Changsha, Hunan, China, in 2021. 

She works at the Shenzhen Health Development Research and Data Management Center. Her res- earch interests include online health information and doctor-patient relationships.
\end{IEEEbiography}

\begin{IEEEbiography}[{\includegraphics[width=1in,height=1.25in,clip,keepaspectratio]{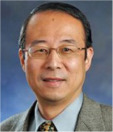}}]{Yi Pan}
is currently a Chair Professor and the Dean of the Faculty of Computer Science and Control Engineering, Shenzhen Institute of Advanced Technologies, Chinese Academy of Sciences and Regents’ Professor Emeritus at Georgia State University. He served as Chair of Computer Science Department at Georgia State University from 2005 to 2020. He has also served as an Interim Associate Dean and Chair of Biology Department during 2013–2017. Dr. Pan joined Georgia State University in 2000, was promoted to full professor in 2004, named a Distinguished University Professor in 2013 and designated a Regents’ Professor (the highest recognition given to a faculty member by the University System of Georgia) in 2015. Dr. Pan received his B.Eng. and M.Eng. degrees in computer engineering from Tsinghua University, China, in 1982 and 1984, respectively, and his Ph.D. degree in computer science from the University of Pittsburgh, USA, in 1991. His profile has been featured as a distinguished alumnus in both Tsinghua Alumni Newsletter and University of Pittsburgh CS Alumni Newsletter. Dr. Pan’s current research interests mainly include bioinformatics and health informatics using big data analytics, cloud computing, and machine learning technologies. Dr. Pan has published more than 450 papers including over 250 journal papers with more than 100 papers published in IEEE/ACM Transactions/Journals. In addition, he has edited/authored 43 books. His work has been cited more than 15000 times based on Google Scholar and his current h-index is 78. Dr. Pan has served as an editor-in-chief or editorial board member for 20 journals including 7 IEEE Transactions. Currently, he is serving as an Associate Editor-in-Chief of IEEE/ACM Transactions on Computational Biology and Bioinformatics. He is the recipient of many awards including one IEEE Transactions Best Paper Award, five IEEE and other international conference or journal Best Paper Awards, 4 IBM Faculty Awards, 2 JSPS Senior Invitation Fellowships, IEEE BIBE Outstanding Achievement Award, IEEE Outstanding Leadership Award, NSF Research Opportunity Award, and AFOSR Summer Faculty Research Fellowship. He has organized numerous international conferences and delivered keynote speeches at over 60 international conferences around the world.
\end{IEEEbiography}
\end{document}